\pdfoutput=1

\documentclass[11pt]{article}

\usepackage{naacl2021}

\usepackage{times}
\usepackage{latexsym}
\usepackage{amsmath}
\usepackage{booktabs}
\usepackage{bbm}
\usepackage{tabularx}
\usepackage{graphicx}
\usepackage{placeins}
\usepackage{footmisc}
\usepackage{stfloats}

\usepackage[T1]{fontenc}

\usepackage[utf8]{inputenc}

\usepackage{microtype}

\newcommand{\method}[0]{\textsc{Fudge}}
\newcommand{\fullname}[0]{Future Discriminators for Generation} 
\newcommand{\pplm}[0]{\textsc{Pplm}}
\newcommand{\cclmspec}[0]{\textsc{Finetune}}
\newcommand{\cclm}[0]{CCLM}
\newcommand{\stytrans}[0]{\textsc{st}}
\newcommand{\vg}[0]{$\mathcal{G}$}
\newcommand{\wdecspec}[0]{\textsc{Wdec}}
\newcommand{\wdec}[0]{WD}
\newcommand{\metric}[0]{$\mathcal{F}$}
\newcommand{\homogenize}[0]{\textsc{FudgeMod}}

\title{\method: Controlled Text Generation With Future Discriminators}

\author{Kevin Yang \\
  UC Berkeley \\
  \texttt{yangk@berkeley.edu} \\\And
  Dan Klein \\
  UC Berkeley \\
  \texttt{klein@berkeley.edu} \\}

\begin{document}
\maketitle
\begin{abstract}

We propose \fullname{} (\method{}), a flexible and modular method for controlled text generation. 
Given a pre-existing model \vg{} for generating text from a distribution of interest, \method{} enables conditioning on a desired attribute $a$ (for example, formality) while requiring access only to \vg{}'s output logits. \method{} learns an attribute predictor operating on a partial sequence, and uses this predictor's outputs to adjust \vg{}'s original probabilities. We show that \method{} models terms corresponding to a Bayesian decomposition of the conditional distribution of \vg{} given attribute $a$. Moreover, \method{} can easily compose predictors for multiple desired attributes. 
We evaluate \method{} on three tasks --- couplet completion in poetry, topic control in language generation, and formality change in machine translation --- and observe gains in all three tasks.

\end{abstract}

\section{Introduction}

Recent advances in large pretrained language models
allow us to generate increasingly realistic text by modeling a distribution $P(X)$ over natural language sequences $X$. The distribution $P(X)$ may be truly unconditional, as is common in language modeling, or it may model $P(X|I)$ conditioned on some input $I$, as in machine translation or summarization.

We are frequently interested in \textit{controlled} text generation, the task of generating text conditioned on an \textit{additional} desirable attribute $a$ which is not already built into $P(X)$. That is, we would like to model $P(X|a)$ (or possibly $P(X|I, a)$; henceforth we will drop $I$ from the notation for simplicity). For example, $P(X)$ may be a pretrained translation model for Spanish inputs $I$ to English outputs $X$, but we may wish to additionally constrain the outputs to possess a new attribute $a$, e.g., formality, which we did not optimize for during training. 

Unfortunately, once we have already obtained an unconditioned $P(X)$ defined as the output distribution of some large generative model \vg{}, it is nontrivial to add conditioning on a new attribute $a$ without either training a new model from scratch or fine-tuning with additional data. Although in principle we can trivially sample from $P(X|a)$ via rejection sampling from $P(X)$, rejection sampling may be highly inefficient in practice. On the other hand, while generating according to attribute $a$, $P(X)$ should be left otherwise intact: in the previous translation formality example, it is pointless to generate formal English outputs if they do not preserve the original Spanish meaning. 




In light of these concerns, we propose 
\fullname{} (\method{}), a flexible and modular method
for modeling $P(X|a)$ which accesses only the output probabilities of the generative model \vg{} which defines $P(X)$.
\method{} learns a binary predictor for whether attribute $a$ will become true in the complete future, based on an incomplete sequence prefix (Sec. \ref{sec:method}). 
Multiplying the output probabilities of this predictor with \vg{}'s original probabilities and then renormalizing yields a model for the desired $P(X|a)$ via Bayes' Rule.

We run experiments on three controlled text generation tasks --- couplet completion in poetry, topic control in language generation, and formality change in machine translation --- showing our method's broad applicability. Additionally, we demonstrate the modularity of \method{} by composing multiple attribute constraints in both the couplet and topic control tasks. In our experiments, we find that \method{} is highly effective at attribute control, outperforming both a baseline which directly fine-tunes \vg{} and also a strong gradient-based method (\pplm{} \cite{dathathri2019plug}). Our code is available at 
 \url{https://github.com/yangkevin2/naacl-2021-fudge-controlled-generation}.
\section{Related Work}


Ideally, a controlled text generation method should efficiently control for $a$ while preserving $P(X)$ as much as possible. Recent work on controlled text generation has greatly advanced our ability to control for a required attribute $a$ flexibly and cheaply, with varying degrees of modification to the original model \vg{} which defines $P(X)$.



One line of work fine-tunes a pretrained model for a desired attribute \cite{ficler2017controlling,yu2017seqgan,ziegler2019fine}. The result is a class-conditional language model (\cclm{}). 
However, it is difficult to isolate the desired attribute from the distribution shift between \vg{} and the fine-tuning dataset \cite{hu2017toward,john2018disentangled,lazaridou2020multi}, i.e., it is nontrivial to preserve the desirable qualities of the $P(X)$ modeled by \vg{}.
One may also need to fine-tune separately for each attribute of interest. 
\textsc{Ctrl} \cite{keskar2019ctrl} partially addresses these issues by providing 55 attribute control codes for a large language model trained from scratch, although this is expensive.
Very recently, \textsc{GeDi} \cite{krause2020gedi} achieves strong performance by using \cclm{} generators as discriminators, though it relies on several heuristics. More broadly, text generation models for style transfer \cite{hu2017toward,lample2018multiple,dai2019style}, summarization \cite{see2017get,gehrmann2018bottom,zaheer2020big}, and machine translation \cite{lample2018phrase,ng2019facebook,lewis2019bart} can also be viewed as \cclm{}'s for different ``attributes.''


A second type of approach instead conditions on a desired attribute by backpropagating gradients, either to directly modify model activations \cite{dathathri2019plug,liu2020data} or to find a trigger string \cite{wallace2019universal,wallace2020imitation}. Such methods often exhibit a high degree of attribute control, and can be used in adversarial attacks \cite{wallace2020imitation}. In fact, \citet{subramani2019can} show that by carefully modifying the latent state, one can cause the base \vg{} to produce arbitrary outputs. 

A third class of methods, referred to as weighted decoding (\wdec{}), assumes access only to $P(X)$ (i.e., \vg{}'s output logits), and operates directly on these logits \cite{ghazvininejad2017hafez,holtzman2018learning,cohn2018pragmatically,shen2019pragmatically}. Compared to other approaches, \wdec{} methods are relatively interpretable in how they obtain $P(X|a)$ from $P(X)$, but prior \wdec{} implementations have been observed to perform poorly in controlled text generation \cite{see2019makes,dathathri2019plug}. 
While \method{} shares a Bayesian motivation with other \wdec{} methods, \method{} follows the Bayesian factorization more closely in implementation (Sec. \ref{sec:method}).
The key distinguishing feature of \method{} is that it models whether attribute $a$ will be true in the \textit{future}, rather than in the \textit{present}. We find that \method{} substantially outperforms previous \wdec{} approaches in our experiments (Sec. \ref{sec:topic_control}).
\begin{figure*}[t!]
    \centering
    \includegraphics[width=0.9\textwidth]{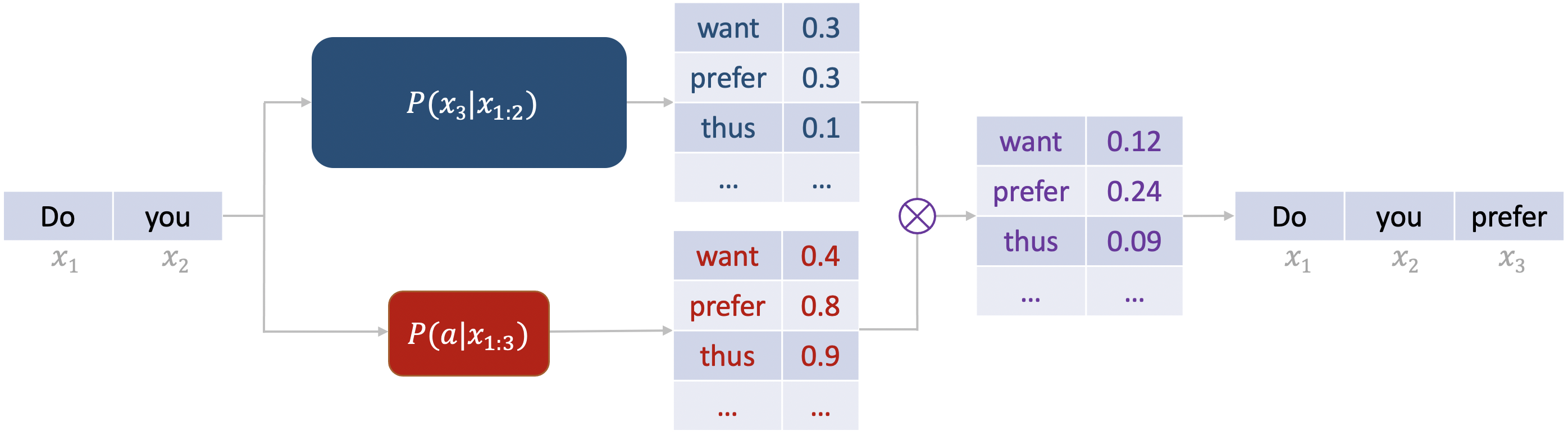}
    \caption{Illustration of one decoding step in \method{}, for an example where the desired attribute $a$ is formality. A large pretrained model \vg{} (dark blue) outputs unconditioned probabilities. Our binary predictor (red) predicts whether the eventual completed sequence will be formal for each possible continuation (computed for each candidate $x_3$, e.g., ``want''; holding $a$ fixed). The probabilities for each $x_3$ are multiplied (purple) and then renormalized to obtain $P(x_3|x_{1:2}, a)$, from which we sample the next token $x_3=$``prefer.''}
    \label{fig:method}
\end{figure*}

\section{Future Discriminators for Generation}\label{sec:method}

We now explain the details of our proposed method, \fullname{} (\method{}), and show that it corresponds to modeling the desired conditional distribution $P(X|a)$. 

For a given language generation task, assume we have an autoregressive model \vg{} (e.g., a large pretrained language model) which models $P(x_i | x_{1:i-1})$ for tokens $x_1 \dots x_i$. Letting $X = x_{1:n}$ denote a completed sequence, \vg{} can sample from $P(X) = P(x_{1:n})$  one token at a time by factoring $P(X)$:

\setlength{\abovedisplayskip}{-5pt}
\setlength{\belowdisplayskip}{5pt}
\begin{align*}
    P(X) = \prod_{i=1}^n P(x_i|x_{1:i-1})
\end{align*}


To condition on attribute $a$, we instead model $P(X|a)$.
This requires a model for $P(x_i | x_{1:i-1}, a)$, modifying the previous factorization:

\begin{align*}
    P(X|a) = \prod_{i=1}^n P(x_i|x_{1:i-1}, a)
\end{align*}

If we model $P(x_i | x_{1:i-1}, a)$ directly, we obtain a class-conditional language model  (\cclm{}). We can learn the \cclm{} by e.g., fine-tuning \vg{} depending on the available data, possibly with some structural modification to \vg{} to accommodate conditioning. 

However, \method{} instead relies on the following Bayesian factorization, exchanging $x_i$ and $a$ conditioned on $x_{1:i-1}$:

\begin{align*}
    P(x_i | x_{1:i-1}, a) \propto P(a | x_{1:i}) P(x_i | x_{1:i-1})
\end{align*}

The second term is exactly the quantity modeled by the base \vg{}. It then suffices to model the first term, $P(a | x_{1:i})$, with a binary classifier $\mathcal{B}$ for the attribute $a$ given a prefix $x_{1:i}$. Intuitively, one can view $\mathcal{B}$ as rescoring or reranking \vg{}'s original hypotheses. 

We emphasize that although $\mathcal{B}$ takes a \textit{prefix} $x_{1:i}$ as input, it predicts whether attribute $a$ will \textit{in the future} be satisfied for the \textit{completed} generation $x_{1:n}$. For instance, suppose we are given a dataset of examples $\{(x_{1:n}, a')\}$ with $a'$ being the values of binary indicators for the desired $a$ (i.e., if $a$ is formality, then $a'$ is 0 or 1 when $x_{1:n}$ is informal or formal respectively). For each training example $(x_{1:n}, a')$, we train our classifier $\mathcal{B}$ using all pairs $(x_{1:i}, a')$; that is, we construct a separate example from each prefix $x_{1:i}$ of $x_{1:n}$. Our approach contrasts with previous methods such as \citet{dathathri2019plug}, which greedily optimize for $a$ on the immediate extension $x_{1:i+1}$. One particular benefit is that \method{} naturally plans for the future: in the example for generating text on the ``space'' topic in Table \ref{topic_examples}, \method{} writes about a ``mysterious ship'' despite ``ship'' itself not being in the given ``space''-topic bag of words, because ``mysterious ship'' easily leads into a mention of one of the targeted ``space'' words (``Earth''). Similarly, in the first couplet completion example in Table \ref{poetry_example}, \method{} needs to rhyme with ``fear'' after exactly ten syllables. After seven syllables, it could reasonably generate the word ``clear,'' but it first generates the adverb ``pretty'' in order to set up the generation of ``clear'' as the tenth syllable.

\method{}'s implementation is shown schematically in Figure \ref{fig:method}, and is quite simple in practice. 
\method{} just needs to learn a $\mathcal{B}$ (red in Figure \ref{fig:method}) sharing tokenization with \vg{} (dark blue). It then converts $\mathcal{B}$'s output into probabilities (red table in Figure \ref{fig:method}), and multiplies with the original output probabilities from \vg{} (dark blue table), to obtain unnormalized probabilities $P(x_i, a|x_{1:i-1})$ (purple table). Finally, renormalizing over the output vocabulary yields the desired distribution $P(x_i|x_{1:i-1}, a)$. In practice, we operate in the log-probability space for numerical stability. 

To improve computational efficiency, we typically 
choose $\mathcal{B}$ to be lightweight relative to \vg{}. We also
consider only the top 200 possibilities for $x_i$ according to \vg{} at each step, as a cheap approximation to the full distribution, and find that this works well in practice.\footnote{See Appendix \ref{appendix:pruning} for ablations on the top-200 pruning.} In each task in Sec. \ref{sec:experiments}, running \method{} on the test set takes no more than 15 minutes on a single Quadro RTX 6000 GPU. 

Finally, as with other controlled generation approaches such as \citet{dathathri2019plug}, it is likely that augmenting \method{} with reranking approaches such as rejection sampling could improve output quality at the cost of compute time, although we do not comprehensively evaluate such extensions in this work. 

\subsection{Advantages and Limitations}

We highlight several additional potential advantages of \method{} compared to directly modeling $P(x_i | x_{1:i-1}, a)$ via e.g., a fine-tuned \cclm{}:
\begin{enumerate}
    \item \method{} requires access only to $P(X)$ (i.e., \vg{}'s output logits) rather than \vg{} itself. 
    \item \vg{} can be freely swapped out for any other model that shares the same tokenization when larger models become available.  
    \item Given multiple conditionally independent attributes with predictors for each, \method{} can easily condition on the combination of these attributes in a modular fashion by summing their output log-probabilities (Sec. \ref{sec:poetry}, \ref{sec:topic_control}).
\end{enumerate}

Unfortunately, like previous methods, \method{} cannot fully guarantee that all outputs possess the desired attribute $a$. 
In \method{}'s case, this is due to the approximation inherent in modeling $P(a | x_{1:i})$, as well as only considering the top 200 possible $x_i$ for computational efficiency. 



\section{Experiments}\label{sec:experiments}

We run experiments on a range of controlled text generation tasks to evaluate the effectiveness of our proposed method: poetry couplet completion (Sec. \ref{sec:poetry}), topic-controlled language generation (Sec. \ref{sec:topic_control}), and machine translation formality change (Sec. \ref{sec:formal_mt}). For each task we discuss the evaluation setup, the specific details of our method and baselines, and finally experimental results. 

\subsection{Poetry Couplet Completion}\label{sec:poetry}


\begin{table}[!htbp]
\small
\begin{tabular}{l}
\toprule
So long as men can breathe or eyes can see,     \\
So long lives this and this gives life to thee.\\
\bottomrule
\end{tabular}
\caption{An example couplet by William Shakespeare. Every second syllable is stressed, following iambic meter, and the last words of each line (see/thee) rhyme.}
\label{shakespeare_example}
\end{table}

We begin with English poetry generation, a task that emphasizes well-formedness, and which has been studied in different forms by many previous works \cite{zhang2014chinese,wang2016chinese,ghazvininejad2016generating,ghazvininejad2017hafez}. Our task here is couplet completion. Given the first line of an iambic pentameter couplet (e.g., Table \ref{shakespeare_example}),
the model must generate a second line which (1) satisfies iambic pentameter, (2) rhymes with the first line, and (3) ends a sentence. The desired attribute $a$ is defined as possessing all three properties, as evaluated by a rule-based checker \metric{} (Appendix \ref{appendix:poetry_checker}).  
Our test set is a collection of prefix lines of couplets, collected from the ending couplet of each of Shakespeare's 154 sonnets.

\textbf{Metrics.} We consider four metrics. 
\begin{enumerate}
    \item \textit{Success}, the fraction of couplet completions with the desired attribute $a$, as checked by \metric{}. This is the main metric.
    \item \textit{Grammaticality}, the probability of grammaticality given by a Roberta-based CoLA grammaticality model \cite{liu2019roberta,warstadt2019neural}, averaged over all outputs.
    \item \textit{Perplexity} of the completion conditioned on the prefix. Following \citet{dathathri2019plug}, since our models use GPT2-Medium \cite{radford2019language} as \vg{}, we evaluate perplexity using GPT \cite{radford2018improving}.\footnote{See Appendix \ref{appendix:perplexity} for other perplexity measurements.}
    \item \textit{Distinctness} of completions, measured as the number of unique unigrams, bigrams, and trigrams across all samples, divided by the total number of words \cite{li2015diversity}. 
\end{enumerate}

At test time, we decode until the model generates ten syllables followed by an end-of-sentence punctuation mark, or after the eleventh syllable (an automatic failure, since iambic pentameter requires exactly ten syllables). 

Overall, because we define $a$ using a rule-based $\mathcal{F}$ which is accessible during training, our formulation of couplet completion is a relatively clean task for evaluating the effectiveness of \method{}. 

\begin{table*}[]
\centering
\begin{tabular}{lcccccc}
\toprule
& \textit{\textbf{Correctness}} & \multicolumn{2}{c}{\textit{\textbf{Text Quality}}} & \multicolumn{3}{c}{\textit{\textbf{Diversity}}}
\\
\cmidrule(lr){2-2} \cmidrule(lr){3-4} \cmidrule(lr){5-7}
  Method          & \textbf{Success} $\uparrow$ & Grammar $\uparrow$ & Perplexity $\downarrow$      & Dist-1 $\uparrow$& Dist-2 $\uparrow$ & Dist-3 $\uparrow$ \\
\cmidrule[\heavyrulewidth](lr){1-1} \cmidrule[\heavyrulewidth](lr){2-2} \cmidrule[\heavyrulewidth](lr){3-4} \cmidrule[\heavyrulewidth](lr){5-7}
\vg{}           & 0 & 0.52   & \phantom{0}44.3 $\pm$ \phantom{0}42.2   & 0.35   & 0.74   & 0.77   \\
\cclmspec{} & 0.21 &	0.44 &	\phantom{0}55.8 $\pm$ \phantom{0}98.3 & 0.35 &	0.74 &	0.78\\
\pplm{}& 0 &	0.54 &	\phantom{0}60.8 $\pm$ \phantom{0}66.1 & 0.40 &	0.78 &	0.78\\
\method{}       & \textbf{0.44} & 0.44   & \phantom{0}70.9 $\pm$ \phantom{0}89.4   & 0.40   & 0.79   & 0.78   \\
\midrule
Shakespeare & \textbf{0.45} & 0.29  & 333.8 $\pm$ 418.9 & 0.44   & 0.81   & 0.79  \\
\bottomrule
\end{tabular}
\caption{Couplet completion results. Success (main metric), grammaticality, perplexity, and distinctness of different methods, tested on 154 prefix lines from Shakespeare sonnets. \method{} substantially outperforms automated baselines on success and maintains high diversity, although quality unsurprisingly suffers compared to the base \vg{} due to the difficult constraint \metric{}. Note Shakespeare's work is often ``incorrect'' due to the narrowness of our metric \metric{};\footref{note1} 
he also scores poorly on text quality because our evaluation models are intended for more modern English.}
\label{tab:results_poetry}
\end{table*}

\subsubsection{Method and Baselines}

\textbf{\method{} Instantiation.} The obvious approach is to learn a predictor for $\mathcal{F}$ directly. However, the three components of $a$ --- meter, rhyme, and sentence-ending --- should be roughly independent. Thus we assume conditional independence, and demonstrate the modularity of \method{} by constructing three separate predictors to be combined at test time:
\begin{enumerate}
    \item $\mathcal{B}_1(x_{1:i})$ takes a text prefix $x_{1:i}$, and predicts whether the completion $x_{1:n}$ of prefix $x_{1:i}$ will be in iambic meter. The model is an LSTM followed by a linear output layer. 
    \item $\mathcal{B}_2(x_{1:i}, t, r)$ takes prefix $x_{1:i}$, the number of syllables $t$ between $x_i$ and $x_n$ for $n \geq i$, and a rhyme sound $r$.\footnote{Two words have the same ``rhyme sound'' $r$ if they rhyme according to the CMU Pronouncing Dictionary \cite{weide1998cmu}.} It predicts whether the completion $x_{1:n}$ has the rhyme sound $r$ at the end of token $x_n$. The model is an LSTM with attention dependent on $t$ and $r$, followed by a shallow feedforward network, and is trained via noise-contrastive estimation \cite{gutmann2010noise}.\footnote{The output logits from $\mathcal{B}_2$ are unnormalized, but this does not affect \method{} after they are added to the output logits of \vg{} and softmaxed for sampling.}
    \item $\mathcal{B}_3(x_{1:i}, t)$ takes prefix $x_{1:i}$ and the number of syllables $t$ between $x_i$ and $x_n$ for $n \geq i$, and predicts whether $x_n$ ends a sentence. The model is an LSTM followed by a shallow feedforward network. 
\end{enumerate} 
The predictors vary in architecture because $\mathcal{B}_2$ and $\mathcal{B}_3$ require inputs other than $x_{1:i}$ --- in truth, they are \textit{families} of related predictors. We find that performance is not overly sensitive to the particulars of the predictor architectures (Appendix \ref{appendix:homogenize}).

To train the discriminators, we sample a dataset of 10 million generations of varied length from GPT2-Medium. From these generations, we sample random subsequences $x_{1:n}$ 
of roughly 10 to 30 syllables 
and truncate $t \leq 10$ ending syllables. These truncations become inputs $x_{1:i}$ to the predictors. For simplicity, we did not balance the class labels for e.g., the iambic predictor during training, although it is likely that doing so would improve performance. 

At test time, we extract $r$ from the given first line of the couplet, and initialize $t = 10$, updating at each step. 
We then modify the output logits of \vg{} by simply adding the log-probabilities from $\mathcal{B}_1$, $\mathcal{B}_2$, and $\mathcal{B}_3$, demonstrating the ease of composing constraints in \method{}.

\textbf{Baselines.} We compare to four baselines.\footnote{A system like Hafez \cite{ghazvininejad2016generating,ghazvininejad2017hafez}, which enforces meter and rhyme at each decoding step using a hard constraint, could achieve perfect success rate. However, this approach relies on the meter and rhyme attributes being ``prefix-checkable'' at the word level: one can guarantee success by simply never selecting a word which immediately violates the constraint. This is often the case for simple rule-based constraints, but not for many other interesting attributes, such as the topic and formality attributes in our subsequent experiments. To preserve generality, \method{} does not rely on this ``prefix-checkable'' property, and neither do our baselines. } 

\begin{enumerate}
    \item \vg{}, the original GPT2-Medium. 
    \item \cclmspec{}, a \cclm{} which finetunes \vg{} on similar inputs to those used for $\mathcal{B}_2$ in \method{}. Since it is not obvious how to compose multiple \cclm{}'s for different attributes, we train a single \cclm{} for all desired properties together. We condition by prefixing the input with (1) whether the last 10 syllables of the original untruncated $x_{1:n}$ are iambic, (2) the rhyme sound at the end of $x_n$, and (3) whether a sentence ends with $x_n$. A special token is inserted 10 syllables from the end of $x_{1:n}$.
    \item \pplm{} \cite{dathathri2019plug}, which uses shallow predictors learned from \vg{}'s top-level hidden layer to modify \vg{}'s states toward increasing probability of the desired attribute via gradient ascent. We decompose the predictors into the same iambic, rhyme sound, and end-of-sentence predictors as for \method{}, inserting an additional hidden layer in the shallow predictor when needed to incorporate additional input (the desired rhyme sound and/or number of syllables until end-of-sentence). 
    \item Shakespeare's original couplet completions.
\end{enumerate}

All non-Shakespeare methods use top-$k$ sampling with $k=10$.

\subsubsection{Results}

Even though our GPT2-Medium-generated training dataset is completely different from the test domain, and contains essentially zero examples of correct couplets, \method{} is able to learn the desired attribute. As shown in Table \ref{tab:results_poetry}, \method{} greatly outperforms all automated baselines in success rate. 

Surprisingly, the \pplm{} baseline achieves zero success. We find that its iambic and rhyme predictors are very poor, so we hypothesize that the relevant information is not easily extractable from the last hidden layer of \vg{}. In contrast, \method{}'s predictors operate directly on the raw text.

Funnily enough, \method{} even matches Shakespeare according to \metric{}, although this is largely due to the narrowness of \metric{} and should not be taken seriously.\footnote{\label{note1} We define \metric{} using somewhat narrow criteria (Appendix \ref{appendix:poetry_checker}), which capture only a subset of what Shakespeare considered to be well-written couplets. The purpose of this task is to evaluate \method{}'s ability to satisfy a difficult well-formedness constraint compared to automated baselines, rather than to perfectly capture the human notion of an iambic pentameter couplet. Thus Shakespeare is marked wrong when he (1) uses archaic pronunciations, (2) uses loose rhymes, (3) elides syllables to fit meter, or (4) uses words missing from the CMU Pronouncing Dictionary. See Appendix \ref{appendix:shakespearefail} for details. Of course, Shakespeare is only included as a whimsical point of reference; our generations obviously do not hold a candle to Shakespeare's originals.}
Similarly, the grammaticality and perplexity metrics are designed for our automated baselines, and thus assign poor scores to Shakespeare's antiquated and flowery style. 

\method{} also maintains relatively fluent generation despite lower grammaticality and perplexity compared to \vg{}.
See Table \ref{poetry_example} for two successful examples. 
Interestingly, \method{} also increases diversity compared to \vg{}, perhaps due to the difficult constraint \metric{} forcing \method{} to use lower-probability regions of the base distribution $P(X)$.

\begin{table}[!htbp]
\small
\begin{tabular}{l}
\toprule
And even thence thou wilt be stol'n, I fear,\\
\textcolor{purple}{for this shall be the end. That's pretty clear.}\\
\midrule
Or, if they sleep, thy picture in my sight\\
\textcolor{purple}{I will be glad to look upon the night.}\\
\bottomrule
\end{tabular}
\caption{Two examples of successful couplet completions (in purple) generated by \method{}.
}
\label{poetry_example}
\end{table}

Finally, it is possible (and trivial) to adjust the conditioning strength in \method{} by multiplying the binary predictors' output logits by a constant. However, this deviates from our Bayesian factorization of $P(X|a)$, and we do not do so.

\subsection{Topic-Controlled Language Generation}\label{sec:topic_control}


\begin{table*}[]
\centering
\begin{tabular}{lcccccc}
\toprule
& \textit{\textbf{On-Topic}} & \multicolumn{2}{c}{\textit{\textbf{Text Quality}}} & \multicolumn{3}{c}{\textit{\textbf{Diversity}}}
\\ \cmidrule(lr){2-2} \cmidrule(lr){3-4} \cmidrule(lr){5-7}
Method &  \textbf{Success} $\uparrow$ & Grammar $\uparrow$ & Perplexity $\downarrow$ & Dist-1 $\uparrow$ & Dist-2 $\uparrow$ & Dist-3 $\uparrow$ \\

\cmidrule[\heavyrulewidth](lr){1-1} \cmidrule[\heavyrulewidth](lr){2-2} \cmidrule[\heavyrulewidth](lr){3-4} \cmidrule[\heavyrulewidth](lr){5-7}
\vg{}                                     & 0.22       &  0.81           & 37.1 $\pm$ 26.9       & 0.35            & 0.78            &0.92            \\
\cclmspec{}                                  & 0.28    & 0.74                & 24.9 $\pm$ 13.7       & 0.29            & 0.70            & 0.88            \\
\wdecspec{}                                   & 0.14    & 0.59                & 33.8 $\pm$ 33.7       & 0.16            & 0.42              & 0.55            \\
\pplm{}                                  & 0.48    & 0.78                & 43.1 $\pm$ 23.7       & 0.35              & 0.78            & 0.92            \\
\method{}                                & \textbf{0.59}    & 0.79                & 40.7 $\pm$ 26.3       & 0.34            & 0.75              & 0.91           \\
\bottomrule
\end{tabular}
\caption{Topic control results. Success (main metric), grammaticality, perplexity, and distinctness for different methods. \cclmspec{} and \wdecspec{} often degenerate into repeating the given bag of words $\mathcal{W}$; this is ill-captured by perplexity, but results in poor grammaticality and distinctness. \method{} substantially outperforms all baselines on success,
including the strong gradient-based \pplm{} baseline, while preserving high quality and diversity.}
\label{tab:results_topic}
\end{table*}

Next, we explore topic control in English language generation. The desired attribute $a$ is to be on-topic for a given topic, such as science or politics. To facilitate comparison with prior work, we largely follow the setup of \pplm{} \cite{dathathri2019plug}: the model is provided an approximation to the topic at test time, in the form of a bag of on-topic words $\mathcal{W}$. The goal is to sample text according to the topic approximated by $\mathcal{W}$, starting from a generic prefix. There are 7 topics (space, politics, military, legal, science, religion, and computers) and 20 prefixes,
and the model generates 3 80-token\footnote{All models and baselines use GPT2 tokenization.} samples from each topic-prefix pair, for a total of 420 generations. 

\textbf{Metrics.} Unfortunately, we cannot easily construct a rule-based $\mathcal{F}$ for being ``on-topic.'' Additionally, use rate of words in $\mathcal{W}$ is a poor metric, because a model can score highly by e.g., simply returning the words in $\mathcal{W}$, without generalizing to the full topic that $\mathcal{W}$ approximates. 
Instead, we adopt a notion of success which requires the model to generalize the bag $\mathcal{W}$ to the full topic. The remaining metrics are measures of quality and diversity.
\begin{enumerate}
    \item \textit{Success}, the average number of distinct words in a heldout bag $\mathcal{W}'$ which appear in the model output. 
    Specifically, for each word in $\mathcal{W}$, we add to $\mathcal{W}'$ the closest GloVe \cite{pennington2014glove} word by cosine similarity, such that the new word does not contain (and is not contained by) any word in $\mathcal{W}$. (This excludes e.g., most plurals.) Usage of distinct words in $\mathcal{W}'$ measures the model's ability to generalize $\mathcal{W}$ to other on-topic words, of which $\mathcal{W}'$ is a non-exhaustive set. This is our main metric. 
    \item \textit{Grammaticality}, identical to the couplet task.
    \item \textit{Perplexity}, identical to the couplet task.
    \item \textit{Distinctness}, defined as in the couplet task. However, it is calculated separately within the 60 generations for each topic, and then averaged over the 7 topics. 
\end{enumerate}

Additionally, following the evaluation procedure of prior work such as \cite{dathathri2019plug}, we run human evaluations via Amazon Mechanical Turk for \method{} against each baseline, comparing topic control and fluency. For each pairwise comparison, we ask 3 workers to evaluate each of 420 paired outputs. Workers were asked to mark which generation is more on topic (first, second, both, or neither), and to rate each generation's fluency on a Likert scale from 1 to 5. We report the average fraction of outputs marked as on-topic as well as the average fluency rating for each method. 

\subsubsection{Method and Baselines}
\textbf{\method{} Instantiation.} Since we model topics as bags of words, \method{} uses a binary predictor $\mathcal{B}(x_{1:i}, w)$ which takes a prefix $x_{1:i}$ and word $w$, and classifies whether $w$ appears in the future $x_{i:n}$ for $n \geq i$. (Since it is desirable to \textit{stay} on topic even after successfully \textit{getting} on topic, we use $x_{i:n}$ rather than $x_{1:n}$.) Training examples $(x_{1:i}, w)$ are sampled from the same dataset of 10 million GPT2-Medium generations used for the couplet  task, and $\mathcal{B}$ is trained using noise-contrastive estimation. $\mathcal{B}$ is a lightweight LSTM-based classifier similar to $\mathcal{B}_2$ from the couplet task.  

At test time, we can compose individual-word constraints if we assume conditional independence between words (although this may be imperfect).
Given a bag of $N$ words $\{w_1 \dots w_N\}$ and prefix $x_{1:i}$, we could condition on all words in the bag appearing in the future by adding all log-probabilities $\log P(w_1|x_{1:i}) \dots \log P(w_N|x_{1:i})$ to \vg{}'s logits. However, topic control does not require every word to appear; perhaps some number $\lambda$ of on-topic words is enough to be ``on-topic.'' Therefore, we model the topic constraint as selecting a random subset of $\lambda$ words from the original bag, and requiring that only those $\lambda$ words all appear. Since each of the $N$ words is selected with probability $\frac{\lambda}{N}$, the quantity we add to the base \vg{} logits is $\frac{\lambda}{N}\sum_{j=1}^N\log P(w_j|x_{1:i})$ in expectation.
In our experiments we use $\lambda=4$, based on a fantasy-topic bag of words used for validation (Appendix \ref{appendix:hyperparams}).

\textbf{Baselines.} We compare to four baselines.
\begin{enumerate}
    \item \vg{}, the original GPT2-Medium. 
    \item \cclmspec{}, which finetunes \vg{} on the same inputs used for \method{}. The future word is given as a prefix for conditioning. At test time, we compute logits for each prefix in the given $\mathcal{W}$ and use the average as the true logits, as an ad hoc way to condition on the full $\mathcal{W}$. 
    \item \wdecspec{}, a simple weighted decoding implementation which greedily considers only the immediate next token when optimizing for $a$. 
    Instead of using $\mathcal{B}$, \wdecspec{} just adds a fixed $\lambda_{\wdecspec{}}$ to the logit for each word in $\mathcal{W}$. Note \wdecspec{} requires $a$ to be well-defined at the token level, so it is not easily transferable to certain tasks (e.g., couplet completion). 
    \item \pplm{} \cite{dathathri2019plug}, which modifies the activations of \vg{} to make the desired bag of words more likely at the immediate next position. We use their method without reranking for fair comparison.
\end{enumerate}

All methods use top-$k$ sampling with $k=10$, following \citet{dathathri2019plug}'s setup.

\subsubsection{Results}

\begin{table}[!htbp]
\addtolength{\tabcolsep}{-3pt}
\centering
\begin{tabular}{lcc}
\toprule
\textbf{Method} & \multicolumn{1}{l}{\textbf{Topic}} & \multicolumn{1}{l}{\textbf{Fluency}} \\
\midrule
\vg{}               & 0.16                               & 4.11                                 \\
\method{}           & \textbf{0.78}                               & \textbf{4.30}                                  \\
\midrule
\cclmspec{}            & 0.24                               & 3.95                                 \\
\method{}           & \textbf{0.76}                               & \textbf{4.22}                                 \\
\midrule
\wdecspec{}              & 0.49                               & 2.50                                  \\
\method{}           & \textbf{0.75}                               & \textbf{4.21}                                 \\
\midrule
\pplm{}            & 0.45                               & 4.05                                 \\
\method{}           & \textbf{0.74}                               & \textbf{4.16}   \\
\bottomrule
\end{tabular}
\caption{Topic control human evaluations, pairwise comparisons. \method{} achieves a substantially higher fraction of on-topic outputs compared to each baseline, in addition to higher average fluency (rated 1 to 5).}
\label{tab:human_eval}
\end{table}

\method{} achieves the highest success by a substantial margin (Table \ref{tab:results_topic}), and outperforms all baselines on human evaluations in both topic relevance and fluency (Table \ref{tab:human_eval}).
\method{} simultaneously preserves high quality and diversity according to automated metrics. Table \ref{topic_examples} shows two examples.

\begin{table}[!htbp]
\small
\begin{tabularx}{\linewidth}{X}
\toprule
\textbf{Space:} The issue focused on the original plot, which was about a mysterious \textcolor{cyan}{ship} that would land on \textcolor{red}{Earth,} and would lead to humanity's first \textcolor{cyan}{interstellar} expedition. The original plan called for humanity to use the \textcolor{red}{spacecraft} to colonize outer \textcolor{red}{space} and build the first city on \textcolor{cyan}{Mars}. But this idea fell by the wayside in the final drafts.\textbackslash{}n\textbackslash{}n"It was just not a very popular idea and it wasn'
\\\midrule
\textbf{Politics:} The issue focused on whether the two \textcolor{cyan}{institutions} were operating within the bounds set by the \textcolor{red}{constitution} and the \textcolor{cyan}{law}.\textbackslash{}n\textbackslash{}nThe \textcolor{red}{Constitutional} \textcolor{red}{Court} said that both governments "have a duty to ensure the integrity of the \textcolor{cyan}{electoral} process and its effective \textcolor{cyan}{administration}, especially in light of the current \textcolor{red}{political} climate that is threatening the functioning of \textcolor{cyan}{elections}"
\\ \bottomrule
\end{tabularx}
\caption{The first output from \method{} when using the prefix ``The issue focused on'' for two topics. We use red to highlight words in the given bag of words $\mathcal{W}$ along with obvious forms (e.g., plurals), and cyan for other on-topic words, including related words not in the heldout bag  $\mathcal{W}'$. 
More examples in Appendix \ref{appendix:additional_examples}.}
\label{topic_examples}
\end{table}

Unsurprisingly, \vg{} performs poorly on success. \wdecspec{} and \cclmspec{} also perform poorly, in success and especially in distinctness. \wdecspec{} frequently degenerates into repeating the given words in the bag $\mathcal{W}$, despite tuning $\lambda_{\wdecspec{}}$ 
(Appendix \ref{appendix:hyperparams}). \cclmspec{} also suffers from repetition, which appears to be the result of distribution shift from fine-tuning. 
Our fine-tuning dataset was built by sampling directly from the original $P(X)$ modeled by \vg{} to mitigate distribution shift, but it is well-known that language model generations are more repetitive than natural language \cite{holtzman2018learning, holtzman2019curious}. We hypothesize that \cclmspec{}, being fine-tuned on language model generations rather than natural language, amplifies this repetitiveness. 
This repetition is reflected in the poor grammaticality for both \cclmspec{} and especially \wdecspec{}. In contrast, \method{} does not touch the original $P(X)$, largely avoiding \cclmspec{}'s distribution shift problem on this task.

Finally, \method{} outperforms the strong gradient-based \pplm{} method, despite requiring access only to \vg{}'s output logits. Non-reliance on gradients means \method{} is also
many times faster than \pplm{}, which takes a few hours compared to \method{}'s 15 minutes for the full set of 420 generations on our hardware. Sometimes we do not even have gradients: for example, gradients are unavailable in the API for GPT3 at time of writing.

\subsection{Machine Translation Formality Change}\label{sec:formal_mt}


Finally, we turn to a somewhat more challenging task, changing formality in machine translation --- specifically, from informal to formal. Given a source sentence written in an informal and conversational style, the goal is to output a translation which is also more formal. We test on the Fisher and CALLHOME Spanish–English Speech Translation Corpus \cite{post2013improved}, a collection of transcribed Spanish conversations with English translations. 
Both the source Spanish and target English are highly informal and disfluent. \citet{salesky2019fluent} augment the Fisher dataset with additional parallel English translations, rewritten to be more fluent (and hence more formal); see Table \ref{trans_example} for an example. Our task is to translate the original informal Spanish to into more formal English.
However, we assume that \citet{salesky2019fluent}'s fluent references are unavailable during training.

\begin{table}[!htbp]
\small
\begin{tabularx}{\linewidth}{X}
\toprule
entonces de verdad sí sí pero entonces tu estudiando para es es digo es más porque es exactamente \\\midrule
Then, if it's business, but then you are a student for a PHD, the Master's is that exactly.       \\\midrule
If it's business, then you are a student for a PhD. The masters is exactly that.      
\\
\bottomrule
\end{tabularx}
\caption{An example from the Fisher dataset. \\\textbf{Top:} The original Spanish transcription. \\\textbf{Middle:} The original English translation. \\\textbf{Bottom:} \citet{salesky2019fluent}'s more fluent version.}
\label{trans_example}
\end{table}

\textbf{Metrics.} The desired attribute $a$ is formality, but we cannot sacrifice the source sentence's meaning. The latter requirement makes generation more constrained than in the couplet and topic tasks, so perplexity and distinctness are less relevant. Instead, we use the following:
\begin{enumerate}
    \item \textit{BLEU Score} \cite{papineni2002bleu}, using two of \citet{salesky2019fluent}'s fluent references per test example. This is our main metric.
    \item \textit{Formality}, the average probability that the model's outputs are formal, according to an evaluator trained on the Family/Relationships domain of the GYAFC formality dataset \cite{rao2018dear}. The evaluator is an LSTM followed by a linear layer. 
\end{enumerate}

\subsubsection{Method and Baselines}
\textbf{\method{} Instantiation.} We assume that the attribute $a$, formality, is conditionally independent from the original conditioning in \vg{}, i.e., the meaning of the Spanish input. \method{} uses a binary predictor $\mathcal{B}(x_{1:n})$ which classifies whether the text starting with prefix $x_{1:n}$ is written in a formal style. $\mathcal{B}$ is an LSTM followed by a linear layer, trained on the Entertainment/Music domain of GYAFC. 

At test time, \method{} directly augments \vg{}'s logits using log-probabilities from $\mathcal{B}$. \vg{} is a pretrained Marian \cite{junczys2018marian} transformer model for Spanish-English. We evaluate both when \vg{} is fine-tuned on the original Fisher training dataset (i.e., using the original targets, not \citet{salesky2019fluent}'s more fluent targets) as well as zero-shot with no fine-tuning, which is challenging due to the highly informal and disfluent text. 

\textbf{Baselines.} We compare to two baselines.
\begin{enumerate}
    \item \vg{}, the original machine translation model.
    \item \vg{} + \stytrans{}, a pipeline consisting of \vg{} followed by a style transfer model. Our style transfer model is T5 \cite{2020t5}, fine-tuned on the same GYAFC Entertainment/Music domain that we used to train $\mathcal{B}$ in \method{}.
\end{enumerate}

Since we do not assume access to \citet{salesky2019fluent}'s more formal targets during training, it is difficult to apply \pplm{} to this task: \pplm{}'s predictor would operate on the pretrained translation model's hidden states, thus requiring a Spanish-English translation dataset with both formal and informal English.\footnote{We nevertheless ran \pplm{} in a somewhat convoluted setup, but found that it performed poorly (Appendix \ref{appendix:pplm_mt}).} We omit \cclmspec{} for the same reason. In contrast, \method{} requires only the original English dataset with formality annotations. 


All methods use greedy decoding.

\subsubsection{Results}

\begin{table}[!htbp]
\addtolength{\tabcolsep}{-3pt}
\begin{tabular}{lcccc}
\toprule
                & \multicolumn{2}{c}{\textit{\textbf{\vg{} (No fine-tune)}}} & \multicolumn{2}{c}{\textit{\textbf{\vg{} (Fine-tune)}}} \\
                \cmidrule(lr){2-3} \cmidrule(lr){4-5}
Method & \textbf{BLEU} $\uparrow$         & Form.  $\uparrow$        & \textbf{BLEU} $\uparrow$     
& Form.  $\uparrow$     \\
\cmidrule[\heavyrulewidth](lr){1-1} \cmidrule[\heavyrulewidth](lr){2-3} \cmidrule[\heavyrulewidth](lr){4-5}
\vg{}               & 16.98                   & 0.45                        & \textbf{22.03}                 & 0.41                       \\
\vg{} + \stytrans{}            & \phantom{0}7.87                    & 0.96                         & \phantom{0}9.63                  & 0.97                      \\
\method{}           & \textbf{17.96}                   & 0.51                        & \textbf{22.18}                 & 0.48                     \\
\bottomrule
\end{tabular}
\caption{Machine translation formality results. BLEU (main metric) and average formality for different methods, with and without fine-tuning \vg{} on the Fisher domain. \method{} increases the formality of translations compared to the base model \vg{} while preserving or increasing BLEU score. Conversely, \vg{} with style transfer overfits to the formality data, resulting in near-perfect formality but losing the original meaning.}
\label{tab:results_trans}
\end{table}


As shown in Table \ref{tab:results_trans}, \method{} increases the formality of outputs compared to \vg{}, even though the test-time formality predictor is trained on a different domain (Family/Relationships, rather than Entertainment/Music). Note that formality unsurprisingly decreases after fine-tuning \vg{}, simply due to the informality of the fine-tuning dataset. As in the couplet task, one could adjust the strength of the formality control in \method{}, although this is unprincipled from the view of modeling $P(X|a)$.

Moreover, while \method{} and \vg{} achieve similar BLEU after fine-tuning \vg{}, \method{} achieves higher BLEU compared to \vg{} when \vg{} is not fine-tuned on the Fisher training set. In the latter case, controlling for formality somewhat remedies the struggles of \vg{} when not fine-tuned on such disfluent text. 

In contrast, the \vg{} + \stytrans{} baseline achieves near-perfect formality but less than half the BLEU of \vg{}, due to the style transfer model overfitting to the GYAFC Entertainment/Music dataset. This is similar to the distribution shift issue that we observed in topic control for \cclmspec{}, an issue which \method{} largely avoids. Nevertheless, there remains substantial room for improvement on this difficult task. 

\begin{table}[!htbp]
\small
\begin{tabular}{ll}
\toprule
Spanish & que era lo que tenía que tienes que hacer \\
\vg{}       & that was what you had to do               \\
\method{}   & That was what you had to do               \\
Reference     & What's there to do?                       \\
\midrule
Spanish & ah en mi en inglaterra por ejemplo        \\
\vg{}       & Ah, in my, in England, for example.       \\
\method{}   & Ah, in England, for example.              \\
Reference     & In England, for example?                  \\
\bottomrule
\end{tabular}
\caption{Example translations by \vg{} (fine-tuned on the Fisher dataset) and \method{} using the same \vg{}. Original Spanish and \citet{salesky2019fluent} references also shown. In this setting, \method{} achieves similar BLEU to \vg{} while increasing formality. While \method{} often simply corrects punctuation or capitalization (top), it also makes more complex adjustments (bottom). More examples in Appendix \ref{appendix:formality_examples}.} 
\label{examples_trans}
\end{table}
\section{Discussion}


\method{} is a principled approach to controlled text generation which models $P(X|a)$ by closely following a Bayesian factorization, thus preserving the base $P(X)$ as much as possible.
\method{} achieves strong performance on a wide range of different tasks: poetry couplet completion, topic control, and informal-to-formal machine translation. Additionally, \method{} can easily compose different attributes in a modular fashion: the meter, rhyme, and end-of-sentence constraints for couplet completion, and the individual words within each topic bag for topic control. In principle, \method{} is applicable to any controlled generation task where we can train discriminators for the desired attribute or attributes.

\section{Ethics of Controlled Text Generation}

We recognize that strong controlled generation methods have the potential to produce harmful outputs and/or misinformation when used adversarially \cite{wallace2019universal,wallace2020imitation}. However, such methods can also be a powerful tool for mitigating harmful biases learned by large pretrained language models \cite{radford2019language,brown2020language}, for example by detoxifying language \cite{dathathri2019plug,krause2020gedi}. Overall, we believe it is still beneficial to continue research into general controlled text generation methods such as \method{}.
\section*{Acknowledgements}

We thank Daniel Fried, David Gaddy, Eric Wallace, Kevin Lin, Nicholas Tomlin, Ruiqi Zhong, and the three anonymous reviewers for their helpful comments and feedback, which aided us in greatly improving the paper. We also thank the authors of \citet{dathathri2019plug} for clarifying our questions about their topic control setup. This work was supported by Berkeley AI Research, DARPA under agreement HR00112020054, and the NSF through a fellowship to the first author. The content does not necessarily reflect the position or the policy of the government, and no official endorsement should be inferred.


\bibliography{anthology,custom}
\bibliographystyle{acl_natbib}

\appendix

\section{Details of $\mathcal{F}$ for Couplet Completion}\label{appendix:poetry_checker}

We provide the full details of the function $\mathcal{F}$ we use to check iambic pentameter,  rhyme, and sentence-ending in our couplet completion task. Note that iambic pentameter consists of two components: iambic meter as well as containing exactly ten syllables.

\begin{enumerate}
    \item \textit{Iambic meter:} Given a phrase, we obtain the sequence of stresses (0 for unstressed, 1 for stressed, 2 for secondary stress) for each word, according to the CMU Pronouncing Dictionary \cite{weide1998cmu}. If any word does not exist in the dictionary (almost never for non-Shakespeare methods) we return False. We treat 2 as ambiguous stress, and additionally change 1 to 2 for any monosyllabic words, i.e. we allow monosyllabic stressed words to be unstressed but not vice versa. Finally, we check that all syllables at even indices (0-indexed) are unstressed or ambiguous, and all syllables at odd indices are stressed or ambiguous. 
    \item \textit{Number of syllables:} We count the number of syllables in each word based on the number of stresses according to the CMU Pronouncing Dictionary. If a word does not exist in the dictionary, we estimate the number of syllables by rounding the number of letters divided by 3 to the nearest integer. 
    \item \textit{Rhyme:} Two words rhyme if and only if they both exist in the CMU Pronouncing Dictionary and are a perfect rhyme according to the dictionary. 
    \item \textit{Sentence-ending:} We check if the output ends with a period, question mark, or exclamation mark. 
\end{enumerate}

Of course, both \method{} and \cclmspec{} will fit to whatever output is given by $\mathcal{F}$. 
The purpose of the couplet task is to check \method{}'s ability to fit a difficult well-formedness constraint. We simply design an $\mathcal{F}$ that corresponds to true iambic pentameter rhymes in most cases. 

\subsection{Shakespeare Evaluation}\label{appendix:shakespearefail}

Shakespeare himself performs somewhat poorly according to \metric{}, which is designed with the automated baselines in mind, not for Shakespeare. (The same is true for our grammaticality and perplexity metrics.) 

One source of error is words which are out-of-vocabulary for the CMU Pronouncing Dictionary. Such words are almost never generated by either \method{} or our automated baselines, but appear in a fifth of Shakespeare's lines, resulting in failures on the iambic meter and syllable checks. 

Nevertheless, most of Shakespeare's ``errors'' are the result of real --- though slight --- deviations from our very strict definitions of meter and rhyme. In particular, he frequently (1) elides syllables to fit meter, and (2) uses loose rhymes; both ``error'' types are likely exacerbated by differences between archaic and modern pronunciations. The example in Table \ref{shakespeare_example_fail} illustrates both types of ``errors.'' Although such deviations are often acceptable to a human, they are difficult to capture in an automatic metric, and we do not allow such deviations in \metric{}. Again, Shakespeare is only included as a whimsical point of reference, and not as a serious baseline to be compared to. 

\begin{table}[!htbp]
\small
\begin{tabular}{l}
\toprule
But here's the joy; my friend and I are one; \\
Sweet flattery! then she loves but me alone.\\
\bottomrule
\end{tabular}
\caption{An example couplet by William Shakespeare, illustrating two common deviations from the narrow definition of correctness we use in \metric{}. For this example to follow iambic meter, one must read ``flattery'' in only two syllables. Moreover, ``one/alone'' is a loose (non-perfect) rhyme, at least in modern English.}
\label{shakespeare_example_fail}
\end{table}


\section{\pplm{} Baseline in Machine Translation}\label{appendix:pplm_mt}

As discussed in the main text, it is difficult to apply \pplm{} in our machine translation setup, in which $P(a|X)$ is learned from an English formality dataset without parallel Spanish. Since $P(X)$ is a Spanish-English translation model, we must obtain hidden states for training \pplm{}’s $P(a|X)$ by first ``backtranslating'' English into Spanish, accessing a second pretrained translator. For this purpose we use a second pretrained Marian transformer from HuggingFace (\url{https://huggingface.co/Helsinki-NLP/opus-mt-en-es}). Additionally, we needed to tune their suggested hyperparameters. 

During evaluation, we observe that \pplm{} makes some reasonable modifications for formality compared to the base $P(X)$, like changing ``hard'' to ``difficult,'' but such improvements are also accompanied by occasional disfluencies and/or repetitions (although such problems plague all methods to some degree). Overall, while \pplm{} achieves similar BLEU to \method{}, it is substantially less formal (Table \ref{tab:mt_pplm}). 

\begin{table}[!htbp]
\centering
\addtolength{\tabcolsep}{-3pt}
\begin{tabular}{lcc}
\toprule
             & \multicolumn{2}{c}{\textit{\textbf{\vg{} (Fine-tune)}}} \\
                \cmidrule(lr){2-3}
Method        & \textbf{BLEU} $\uparrow$     
& Form.  $\uparrow$     \\
\cmidrule[\heavyrulewidth](lr){1-1} \cmidrule[\heavyrulewidth](lr){2-3}

\pplm{} &  21.94 & 0.40 \\
\method{}                          & 22.18                 & 0.48                     \\
\bottomrule
\end{tabular}
\caption{\pplm{} baseline in machine translation formality on the fine-tuned \vg{}.}
\label{tab:mt_pplm}
\end{table}

\section{Hyperparameter Choices}\label{appendix:hyperparams}

\method{} has essentially one hyperparameter in our topic control task, $\lambda$, which controls the strength of conditioning and corresponds to the number of words in the bag which should appear in the future. 

To choose $\lambda$ in topic control, we used a separate validation bag of words (on the topic of fantasy; Appendix \ref{appendix:fantasy}) to select a reasonable $\lambda$ for our main paper experiments ($\lambda = 4$). Unlike in the main paper where we use heldout bags $\mathcal{W}'$ to measure success, during validation we simply use the original bag. We use a set of 60 generations, considering values ranging from 1 to 6 (Table \ref{tab:lambda_opt}), although the result may be somewhat noisy. Of course, different choices of $\lambda$ result in different tradeoffs (Appendix \ref{appendix:lambda}).

\begin{table*}[]
\centering
\begin{tabular}{lcccccc}
\toprule
& \textit{\textbf{On-Topic}} & \multicolumn{2}{c}{\textit{\textbf{Text Quality}}} & \multicolumn{3}{c}{\textit{\textbf{Diversity}}}
\\ \cmidrule(lr){2-2} \cmidrule(lr){3-4} \cmidrule(lr){5-7}
Method &  \textbf{Success} $\uparrow$ & Grammar $\uparrow$ & Perplexity $\downarrow$ & Dist-1 $\uparrow$ & Dist-2 $\uparrow$ & Dist-3 $\uparrow$ \\
\cmidrule[\heavyrulewidth](lr){1-1} \cmidrule[\heavyrulewidth](lr){2-2} \cmidrule[\heavyrulewidth](lr){3-4} \cmidrule[\heavyrulewidth](lr){5-7}
\method{}, $\lambda=1$               & 0.05 &  0.80                  & 38.3 $\pm$ 32.6       & 0.36           & 0.78           & 0.92           \\
\method{}, $\lambda=2$               & 0.10 &  0.76                   & 31.1 $\pm$ 17.1       & 0.35           & 0.75           & 0.91           \\
\method{}, $\lambda=4$               & 0.28 &  0.76                  & 40.1 $\pm$ 27.6       & 0.37           & 0.77           & 0.92           \\
\method{}, $\lambda=6$               & 0.30 &  0.72                   & 46.9 $\pm$ 29.9       & 0.38           & 0.77           & 0.91          \\\bottomrule
\end{tabular}
\caption{Results from 60 samples for \method{} with different $\lambda$ on topic control for a validation fantasy-topic bag of words. Note that during validation only, success directly measures use rate of words in the given bag $\mathcal{W}$, not a heldout bag $\mathcal{W}'$ as in the main paper.}
\label{tab:lambda_opt}
\end{table*}

\begin{table*}[]
\centering
\begin{tabular}{lcccccc}
\toprule
& \textit{\textbf{On-Topic}} & \multicolumn{2}{c}{\textit{\textbf{Text Quality}}} & \multicolumn{3}{c}{\textit{\textbf{Diversity}}}
\\ \cmidrule(lr){2-2} \cmidrule(lr){3-4} \cmidrule(lr){5-7}
Method &  \textbf{Success} $\uparrow$ & Grammar $\uparrow$ & Perplexity $\downarrow$ & Dist-1 $\uparrow$ & Dist-2 $\uparrow$ & Dist-3 $\uparrow$ \\
\cmidrule[\heavyrulewidth](lr){1-1} \cmidrule[\heavyrulewidth](lr){2-2} \cmidrule[\heavyrulewidth](lr){3-4} \cmidrule[\heavyrulewidth](lr){5-7}
\wdecspec{}, $\lambda_{\wdecspec{}}=1$              & 0.02   &  0.83                & 34.5 $\pm$ 23.5       & 0.36            & 0.78            & 0.91            \\
\wdecspec{}, $\lambda_{\wdecspec{}}=2$               & 0.02  &   0.83                & 34.9 $\pm$ 23.8       & 0.36            & 0.78            & 0.91            \\
\wdecspec{}, $\lambda_{\wdecspec{}}=4$               & 0.57  &   0.79                & 34.7 $\pm$ 23.6       & 0.33            & 0.74            & 0.86            \\
\wdecspec{}, $\lambda_{\wdecspec{}}=8$               & 1.90  &   0.47                 & 14.5 $\pm$ 19.2       & 0.04            & 0.09            & 0.12            \\
\wdecspec{}, $\lambda_{\wdecspec{}}=16$              & 2.32  &   0.40                & \phantom{0}8.4 $\pm$ \phantom{0}9.6         & 0.01            & 0.04            & 0.06            \\
\wdecspec{}, $\lambda_{\wdecspec{}}=32$              & 2.35  &   0.41                & \phantom{0}7.5 $\pm$ \phantom{0}9.1         & 0.01            & 0.04            & 0.06  \\\bottomrule       
\end{tabular}
\caption{Results from 60 samples for \wdecspec{} with different $\lambda_{\wdecspec{}}$ on topic control for a validation fantasy-topic bag of words. Note that during validation only, success directly measures use rate of words in the given bag $\mathcal{W}$, not a heldout bag $\mathcal{W}'$ as in the main paper.}
\label{tab:lambdawd_opt}
\end{table*}

We also optimized the conditioning strength $\lambda_{\wdecspec{}}$ for the \wdecspec{} baseline on the same fantasy bag of words, considering values ranging from 1 to 32. We selected the only value (4) which achieved reasonable success without a total collapse in diversity (Table \ref{tab:lambdawd_opt}), but diversity still collapsed when tested on our seven main test bags of words. 

We do not optimize any model hyperparameters in the couplet completion and informal-to-formal translation tasks. LSTM's and feedforward networks are 3 layers (including the output layer of dimension 1) and 300-dimensional unless otherwise specified. They are bidirectional (150-dimensional in each direction) for the couplet rhyme predictor and the topic control future words predictor, and otherwise unidirectional. Attention mechanisms use key-query-value attention. For the rhyme and future words predictors the output hidden state is multiplied element-wise by the embedding of the rhyme sound or future word, then concatenated to the embeddings, before the final feedforward network. Since a selling point of our method is the lightweight process of constructing and training predictors, noise-contrastive estimation is a natural choice for the rhyme and future word predictors: we avoid softmaxing over the output dimension during training. (This is primarily relevant for the future word predictor, as the number of distinct rhyme sounds is not too large, but we use noise-contrastive estimation for both for consistency's sake.)

For the \pplm{} baseline, we used step size 0.01 for both couplet completion and MT after tuning, and kept their other hyperparameters fixed. For topic control we simply evaluated their provided generations instead of rerunning their model. 

\begin{table*}[]
\centering
\begin{tabular}{lcccccc}
\toprule
& \textit{\textbf{Correctness}} & \multicolumn{2}{c}{\textit{\textbf{Text Quality}}} & \multicolumn{3}{c}{\textit{\textbf{Diversity}}}
\\
\cmidrule(lr){2-2} \cmidrule(lr){3-4} \cmidrule(lr){5-7}
  Method          & \textbf{Success} $\uparrow$ & Grammar $\uparrow$ & Perplexity $\downarrow$      & Dist-1 $\uparrow$& Dist-2 $\uparrow$ & Dist-3 $\uparrow$ \\
\cmidrule[\heavyrulewidth](lr){1-1} \cmidrule[\heavyrulewidth](lr){2-2} \cmidrule[\heavyrulewidth](lr){3-4} \cmidrule[\heavyrulewidth](lr){5-7}
\method{}       & 0.44 & 0.44   & \phantom{0}70.9 $\pm$ \phantom{0}89.4   & 0.40   & 0.79   & 0.78   \\
\homogenize{}& 0.39 &	0.43 &	\phantom{0}72.1 $\pm$ \phantom{0}66.3 & 0.41 &	0.79 &	0.77\\
\bottomrule
\end{tabular}
\caption{Ablation of \method{} with a modified predictor architecture on couplet completion.}
\label{tab:homogenize_poetry}
\end{table*}

\begin{table*}[]
\centering
\begin{tabular}{lcccccc}
\toprule
& \textit{\textbf{On-Topic}} & \multicolumn{2}{c}{\textit{\textbf{Text Quality}}} & \multicolumn{3}{c}{\textit{\textbf{Diversity}}}
\\ \cmidrule(lr){2-2} \cmidrule(lr){3-4} \cmidrule(lr){5-7}
Method &  \textbf{Success} $\uparrow$ & Grammar $\uparrow$ & Perplexity $\downarrow$ & Dist-1 $\uparrow$ & Dist-2 $\uparrow$ & Dist-3 $\uparrow$ \\

\cmidrule[\heavyrulewidth](lr){1-1} \cmidrule[\heavyrulewidth](lr){2-2} \cmidrule[\heavyrulewidth](lr){3-4} \cmidrule[\heavyrulewidth](lr){5-7}
\method{}                                & 0.59    & 0.79                & 40.7 $\pm$ 26.3       & 0.34            & 0.75              & 0.91           \\
\homogenize{}                                  & 0.62    & 0.77                & 47.8 $\pm$ 51.3       & 0.33              & 0.73            & 0.88            \\
\bottomrule
\end{tabular}
\caption{Ablation of \method{} with a modified predictor architecture on topic control.}
\label{tab:homogenize_topic}
\end{table*}

\section{Ablations on Predictor Architectures}\label{appendix:homogenize}

Some variation in predictor architectures is necessary due to the diversity of our tasks (as evidenced by the difficulties in adapting PPLM). Specifically, while our core predictor architecture is word embeddings followed by LSTM and output layer, task-specific architectures vary because some ``predictors'' are actually families of related predictors. We model such families as a single predictor taking additional input (e.g., rhyme sound in poetry); this is needed in our poetry and topic tasks. 

On these two tasks, we provide ablations with more homogenized predictors: additional inputs are simply embedded and concatenated to each input word embedding. The difference is relatively small in both cases (Tables \ref{tab:homogenize_poetry} and \ref{tab:homogenize_topic}). \homogenize{} indicates the ablated version of \method{}.

\begin{table*}[]
\centering
\begin{tabular}{lccc}
\toprule
\textbf{Method} & \textbf{GPT}           & \textbf{TFXL}    & \textbf{GPT-Shakespeare}           \\
\midrule
\vg{}      & \phantom{0}44.3 $\pm$ \phantom{0}42.2 & \phantom{00}84.8 $\pm$ \phantom{0}111.0  &  \phantom{0}72.9 $\pm$ \phantom{0}62.0 \\
\cclmspec{}   & \phantom{0}55.8 $\pm$ \phantom{0}98.3 & \phantom{00}76.1 $\pm$ \phantom{00}64.6  &  \phantom{0}69.0 $\pm$ \phantom{0}92.5  \\
\pplm{}   & \phantom{0}60.8 $\pm$ \phantom{0}66.1 & \phantom{0}111.5 $\pm$ \phantom{0}150.4  &  120.0 $\pm$ 243.8  \\
\method{}  & \phantom{0}70.9 $\pm$ \phantom{0}89.4 & \phantom{0}137.5 $\pm$ \phantom{0}170.9 & \phantom{0}96.2 $\pm$ 117.3 \\  
\midrule
Shakespeare   & 333.8 $\pm$ 418.9 & 1879.5 $\pm$ 6088.1 & 195.1 $\pm$ 228.9    \\
\bottomrule
\end{tabular}
\caption{Different perplexity measurements on couplet completion, using GPT, Transformer-XL (TFXL), and GPT fine-tuned with Shakespearean language (GPT-Shakespeare). Main paper results use GPT.}
\label{tab:perplexity_poetry}
\end{table*}

\begin{table*}[]
\centering
\begin{tabular}{lcc}
\toprule
\textbf{Method} & \textbf{GPT}           & \textbf{TFXL}               \\
\midrule
\vg{}      & 37.1 $\pm$ 26.9 & \phantom{00}34.1 $\pm$ \phantom{000}25.2      \\
\cclmspec{}   & 24.9 $\pm$ 13.7 & \phantom{00}27.7 $\pm$ \phantom{000}15.6      \\
\wdecspec{}     & 33.8 $\pm$ 33.7 & 7802.4 $\pm$ 29942.6 \\
\pplm{}   & 43.1 $\pm$ 23.7 & \phantom{00}38.7 $\pm$ \phantom{000}21.0      \\
\method{}  & 40.7 $\pm$ 26.3 & \phantom{00}42.8 $\pm$ \phantom{000}46.9  \\   
\bottomrule
\end{tabular}
\caption{Different perplexity measurements on topic control, using GPT and Transformer-XL (TFXL). Main paper results use GPT.}
\label{tab:perplexity_topic}
\end{table*}

\section{Alternative Perplexity Measurements}\label{appendix:perplexity}

On the couplet completion task, we additionally measure perplexity using Transformer-XL \cite{dai2019transformer} and using a GPT model fine-tuned on Shakespearean language as generated by \cite{lau2018deep}. We measure using Transformer-XL on the topic control task as well. Relative perplexities between most models remain largely similar when switching between GPT and Transformer-XL, with a few exceptions. Compared to the base GPT, Shakespeare's perplexity naturally decreases while other models' perplexities increase when measured with Shakespeare-finetuned GPT. The highly repetitive and disfluent \wdecspec{} baseline is rightly punished for this behavior when measured by Transformer-XL. \pplm{} also obtains slightly lower perplexity than \method{} on topic control when measured by Transformer-XL. Full results in Tables \ref{tab:perplexity_poetry} and \ref{tab:homogenize_topic}.


\section{Statistical Significance}\label{appendix:significance}

In couplet completion, \method{} outperforms the strongest automated baseline (\cclmspec{}) on success rate with $p < 0.0001$ on a McNemar test, pairing the generations for each Shakespeare prefix. 

In topic control, \method{} outperforms the strongest automated baseline \pplm{} with $p = 0.04$ using a Wilcoxon matched pairs test, pairing the generations for topic-prefix combinations. 

In translation formality, \method{}'s generations are more formal than those of the base \vg{} with $p < 0.0001$ according to a paired t-test. 


\begin{table}[!htbp]
\small
\begin{tabularx}{\linewidth}{X}
\toprule
\textbf{Space:} The issue focused on the new, higher level of control that \textcolor{cyan}{NASA} had in the \textcolor{red}{space} \textcolor{cyan}{shuttle} program.\textbackslash{}n\textbackslash{}n"The question of how far the U.S. government can extend its jurisdiction in \textcolor{red}{space} was raised," Mr. Smith said.\textbackslash{}n\textbackslash{}n\textcolor{cyan}{NASA's} role has become increasingly important in the 21st century in part because of the growth in \textcolor{red}{space} activities. The \textcolor{red}{space} \textcolor{cyan}{shuttle} program began in 1977 with \\ \midrule
\textbf{Politics:} The issue focused on how much \textcolor{red}{power} each company was willing to use in response to the request.\textbackslash{}n\textbackslash{}nAccording to the complaint, Comcast has not been forthcoming with any data, such as how often it uses the technology, and what it has paid for it, in \textcolor{red}{order} to meet the \textcolor{cyan}{FCC's} mandate to make its own data more accessible.\textbackslash{}n\textbackslash{}nAnd, according to the suit, the company also                         \\ \midrule
\textbf{Military:} The issue focused on the use of \textcolor{red}{force} by the \textcolor{red}{armed} \textcolor{red}{forces} and \textcolor{cyan}{police}, as well as the use of \textcolor{cyan}{lethal} \textcolor{red}{force} by \textcolor{cyan}{civilians}.\textbackslash{}n\textbackslash{}nThe bill would require that a \textcolor{red}{shooting} occur "with reasonable care," meaning a \textcolor{red}{shot} was "justified" under the circumstances of the case and not in retaliation for an act of \textcolor{cyan}{violence,} and that a \textcolor{red}{shooting} was "necessary for the \textcolor{cyan}{safety} of the \textcolor{red}{officer} or the         \\\bottomrule                                     
\end{tabularx}
\caption{The first generation by \method{} using $\lambda=2$ on the space, politics, and military topics given the prefix ``The issue focused on.'' Words in the given bag are highlighted in red, and other related words in cyan. }
\label{example:lam2}
\end{table}

\section{Effect of Varying Topic Control Strength}\label{appendix:lambda}

Although we use $\lambda=4$ for \method{} in our main paper experiments for topic control, we experiment here with varying the conditioning strength. Specifically, we experiment with $\lambda=2$ and $\lambda=8$. The conditioning is unsurprisingly stronger as $\lambda$ increases, as shown quantitatively in Table \ref{tab:lambda}, although the perplexity increases as well. 

We also provide some example generations for $\lambda=2$ and $\lambda=8$ in Tables \ref{example:lam2} and \ref{example:lam8}, for the same prompts and topics as in Table \ref{topic_examples} for $\lambda=4$ in the main text. The $\lambda=8$ generations remain mostly fluent and interesting, despite their worse grammaticality and perplexity.

\section{Effect of Varying Candidate Pruning}\label{appendix:pruning}

For computational efficiency, we only feed the top 200 candidates returned by \vg{} into \method{}'s predictor when predicting each next token. Here, we ablate on this number in our topic control setting, testing 100 and 400 (Table \ref{tab:candidate_pruning}).


\begin{table*}[]
\centering
\begin{tabular}{lcccccc}
\toprule
& \textit{\textbf{On-Topic}} & \multicolumn{2}{c}{\textit{\textbf{Text Quality}}} & \multicolumn{3}{c}{\textit{\textbf{Diversity}}}
\\ \cmidrule(lr){2-2} \cmidrule(lr){3-4} \cmidrule(lr){5-7}
Method &  \textbf{Success} $\uparrow$ & Grammar $\uparrow$ & Perplexity $\downarrow$ & Dist-1 $\uparrow$ & Dist-2 $\uparrow$ & Dist-3 $\uparrow$ \\
\cmidrule[\heavyrulewidth](lr){1-1} \cmidrule[\heavyrulewidth](lr){2-2} \cmidrule[\heavyrulewidth](lr){3-4} \cmidrule[\heavyrulewidth](lr){5-7}
\method{}, $\lambda=2$                                & 0.49  &  0.79                 & 38.6 $\pm$ 24.1       & 0.36              & 0.76              & 0.91              \\
\method{}, $\lambda=4$                                   & 0.59  &  0.79                 & 40.7 $\pm$ 26.3       & 0.34              & 0.75              & 0.91              \\
\method{}, $\lambda=8$                                & 0.76   &  0.74                & 56.5 $\pm$ 39.0       & 0.35              & 0.75              & 0.90   \\\bottomrule
\end{tabular}
\caption{\method{} results for different values of $\lambda$ on the main 7 topics and 20 prefixes. Success and perplexity both increase as the conditioning strength $\lambda$ increases. Our main paper experiments use $\lambda=4$.}
\label{tab:lambda}
\end{table*}

\begin{table*}[!htbp]
\centering
\begin{tabular}{lcccccc}
\toprule
& \textit{\textbf{On-Topic}} & \multicolumn{2}{c}{\textit{\textbf{Text Quality}}} & \multicolumn{3}{c}{\textit{\textbf{Diversity}}}
\\ \cmidrule(lr){2-2} \cmidrule(lr){3-4} \cmidrule(lr){5-7}
Method &  \textbf{Success} $\uparrow$ & Grammar $\uparrow$ & Perplexity $\downarrow$ & Dist-1 $\uparrow$ & Dist-2 $\uparrow$ & Dist-3 $\uparrow$ \\
\cmidrule[\heavyrulewidth](lr){1-1} \cmidrule[\heavyrulewidth](lr){2-2} \cmidrule[\heavyrulewidth](lr){3-4} \cmidrule[\heavyrulewidth](lr){5-7}
\method{}, $100$                                & 0.54  &  0.77                 & 48.9 $\pm$ 37.9       & 0.36              & 0.75              & 0.91              \\
\method{}, $200$                                   & 0.59  &  0.79                 & 40.7 $\pm$ 26.3       & 0.34              & 0.75              & 0.91              \\
\method{}, $400$                                & 0.61   &  0.77                & 49.4 $\pm$ 56.9       & 0.35              & 0.76              & 0.91   \\\bottomrule
\end{tabular}
\caption{\method{} results for different numbers of candidates fed through \method{}'s predictor. Main paper results use 200. }
\label{tab:candidate_pruning}
\end{table*}

\begin{table}[!htbp]
\small
\begin{tabularx}{\linewidth}{X}
\toprule
\textbf{Space:} The issue focused on the size of \textcolor{cyan}{NASA's} \textcolor{red}{satellite} \textcolor{cyan}{telescope} that is being used to observe the \textcolor{cyan}{universe}.\textbackslash{}n\textbackslash{}nThe \textcolor{cyan}{telescope} will be the \textcolor{cyan}{world's} largest when it is completed in 2022.\textbackslash{}n\textbackslash{}nThe US \textcolor{red}{space} agency wants to put the \textcolor{cyan}{telescope} into a new \textcolor{red}{orbit} around the \textcolor{red}{planet}.\textbackslash{}n\textbackslash{}nThe \textcolor{cyan}{Hubble} \textcolor{red}{Space} \textcolor{cyan}{Telescope} \textcolor{red}{orbits} in an elliptical \textcolor{red}{orbit,} which puts the \textcolor{cyan}{telescope} into a "cross-path" \\\midrule
\textbf{Politics:} The issue focused on the \textcolor{red}{power,} \textcolor{cyan}{independence} and \textcolor{cyan}{independence} of the \textcolor{cyan}{federal} \textcolor{cyan}{judiciary}. In its ruling, the three-\textcolor{cyan}{judge} \textcolor{cyan}{"progressive"} panel of the 10th Circuit of the \textcolor{cyan}{U.S.} \textcolor{red}{Court} of Appeals for the 10th Circuit found that the "\textcolor{cyan}{Supreme} \textcolor{red}{Court} lacks the requisite \textcolor{red}{power} to make decisions on the \textcolor{red}{constitutionality} of any particular \textcolor{cyan}{federal} \textcolor{cyan}{regulation}, including the \textcolor{red}{power} to make the final determination                                                                                               \\\midrule
\textbf{Military:} The issue focused on the \textcolor{red}{military} wing of the \textcolor{cyan}{U.S.} \textcolor{red}{Navy} that manages \textcolor{cyan}{ships} to the surface of a seabed, the \textcolor{cyan}{USS} Ponce, which is carrying a guided-\textcolor{red}{missile} \textcolor{red}{destroyer.} A small accident with the \textcolor{red}{weapon's} guidance system forced the \textcolor{red}{carrier's} \textcolor{cyan}{"supergun"} system to be deactivated, resulting in a loss of a "\textcolor{red}{kill} \textcolor{cyan}{ship}"role from the \textcolor{red}{fleet.} The                               \\\bottomrule                                                                                              
\end{tabularx}
\caption{The first generation by \method{} using $\lambda=8$ on the space, politics, and military topics given the prefix ``The issue focused on.'' Words in the given bag are highlighted in red, and other related words in cyan. }
\label{example:lam8}
\end{table}

\section{Additional Couplet Completion Examples}\label{appendix:poetry_examples}

We provide some additional examples of \method{} and baselines on our couplet completion task in Table \ref{tab:poetry_additional_examples}.

\begin{table}[!htbp]
\small
\begin{tabularx}{\linewidth}{X}
\toprule
\textbf{Prefix:} Pity the world, or else this glutton be,       \\
\textbf{\vg{}:}      (1) to see that you are not to be ashamed.      \\
\textbf{\cclm{}:} or else it is a human being. He.\\
\textbf{\pplm{}:}    ..." He continued, "I have seen, and heard of the              \\
\textbf{\method{}:}     and so I do; and so I do, as be.                  \\
\textbf{Shakespeare:}    To eat the world's due, by the grave and thee.                 \\
\midrule
\textbf{Prefix:} But flowers distill'd though they with winter meet,       \\
\textbf{\vg{}:}      and with their sweet, sweet love; The snow-white rose      \\
\textbf{\cclm{}:} the flowers are not quite so sweet or sweet.\\
\textbf{\pplm{}:}    'Twas a great feast, and merry company all              \\
\textbf{\method{}:}     (I will be here to see her) let my sweet.                  \\
\textbf{Shakespeare:}    Leese but their show; their substance still lives sweet. \\   
\midrule
\textbf{Prefix:}  And nothing 'gainst Time's scythe can make defence       \\
\textbf{\vg{}:}      A better foe to be slain, than Death, and all      \\
\textbf{\cclm{}:} a better weapon than it is. That defence\\
\textbf{\pplm{}:}    I am so glad that I have found a new home              \\
\textbf{\method{}:}     worth more. That's how much time is on the fence.                  \\
\textbf{Shakespeare:}    Save breed, to brave him when he takes thee hence. \\ 
\bottomrule
\end{tabularx}
\caption{Additional example couplet completions, comparing \vg{}, \cclmspec{}, \pplm{}, \method{}, and Shakespeare. ``Prefix'' indicates the first line of the couplet which generations are conditioned on. } 
\label{tab:poetry_additional_examples}
\end{table}

We also show some unsuccessful examples for \method{} in \ref{tab:poetry_additional_examples_fail}. Overall, we find that most errors are due to the rhyme and ten-syllable end of sentence constraints, or due to Shakespeare's prefix ending in a word not in the CMU Pronouncing Dictionary (e.g., ``prognosticate'' in the table). \method{} also sometimes overgenerates punctuation at the end of a sentence.

\begin{table}[!htbp]
\small
\begin{tabularx}{\linewidth}{X}
\toprule
\textbf{Prefix:} Make thee another self, for love of me,       \\
\textbf{\method{}:}     and for the joy of thee, and for love's,.... I,.                  \\
\textbf{Shakespeare:}    That beauty still may live in thine or thee.                 \\
\midrule
\textbf{Prefix:} Or else of thee this I prognosticate:       \\
\textbf{\method{}:}     I, O, Lord Jesus Christ, my Beloved; I am                  \\
\textbf{Shakespeare:}    Thy end is truth's and beauty's doom and date. \\   
\midrule
\textbf{Prefix:}  But were some child of yours alive that time,       \\
\textbf{\method{}:}     would you still have loved that child for that? I'm...                  \\
\textbf{Shakespeare:}    You should live twice; in it and in my rhyme. \\ 
\bottomrule
\end{tabularx}
\caption{Additional example couplet completions where \method{} is unsuccessful. In the first, \method{} fails to rhyme; in the second, ``prognosticate'' is not in the CMU Pronouncing Dictionary; in the third, \method{} gets the syllable count wrong.} 
\label{tab:poetry_additional_examples_fail}
\end{table}

\section{Additional Topic Control Examples}\label{appendix:additional_examples}

In Tables \ref{examples_moreimportantly}, \ref{examples_ithasbeenshown}, and \ref{examples_toreview} we show additional example generations by our method using the same hyperparameter setting as in the main paper, $\lambda = 4$. Specifically, we provide the first generation by \method{} for 3 separate prefixes for each of the 7 topics. Virtually all examples are clearly on topic, while avoiding repetitiveness. 

Additionally, we provide example generations from \vg{}, \cclmspec{} and \wdecspec{} in Tables \ref{tab:g_topic_examples}, \ref{tab:cclm_topic_examples}, and \ref{tab:wdec_topic_examples} respectively. For \pplm{} we refer the reader to the examples in the main paper and appendices of \citet{dathathri2019plug}'s original work.

\begin{table}[]
\small
\begin{tabularx}{\linewidth}{X}
\toprule
\textbf{Space:} More importantly, the E4E-R-E-S-T report finds, the greatest threat to \textcolor{red}{Earth's} existence comes from the human race's inability to adapt: "It is clear from E4E's analysis that a lack of knowledge about how to manage an expanding \textcolor{cyan}{world} and to adapt to changing climatic conditions poses a serious challenge to our ability to sustain life on \textcolor{red}{planet} \textcolor{red}{Earth.}                                                                                \\\midrule
\textbf{Politics:} More importantly, in an effort to preserve the historical integrity of the \textcolor{red}{state's} \textcolor{cyan}{judicial} \textcolor{cyan}{system}, the \textcolor{red}{state} also needs to ensure its integrity within the larger \textcolor{cyan}{American} \textcolor{red}{political} \textcolor{cyan}{system} through fair, transparent, and competitive \textcolor{cyan}{elections.} In other words, a \textcolor{cyan}{system} based upon \textcolor{cyan}{meritocracy} and \textcolor{red}{equality} for all \textcolor{cyan}{candidates, voters, candidates and parties.}\textbackslash{}n\textbackslash{}nThe \textcolor{cyan}{American people} have a \textcolor{cyan}{right} to know whether the current \textcolor{cyan}{system} for \textcolor{cyan}{electing} \textcolor{red}{state}      \\\midrule
\textbf{Military:} More importantly, the \textcolor{red}{military} has a great deal of leverage with its own \textcolor{red}{soldiers} and their superiors, and its willingness to use that leverage to \textcolor{red}{force} an immediate end to this practice of indefinite \textcolor{cyan}{detention} and indefinite \textcolor{cyan}{imprisonment} was demonstrated by the recent court \textcolor{cyan}{order} issued by the US District Court for the District of Columbia, which requires the release of an Iraqi-American held in an indefinite \textcolor{red}{military} \textcolor{cyan}{detentions facility} and a \textcolor{cyan}{detainee}                        \\\midrule
\textbf{Legal:} More importantly, in an effort to make the \textcolor{red}{case} that the \textcolor{red}{law} was needed because of its effects on the \textcolor{red}{state's} economy, the \textcolor{red}{law's} supporters claimed that the "\textcolor{cyan}{death penalty} was necessary to protect public safety." (The \textcolor{red}{argument} that the \textcolor{cyan}{punishment} was needed because it was needed to prevent certain \textcolor{red}{crimes} was rejected by the \textcolor{red}{Court.)} The state's \textcolor{red}{argument} was that the \textcolor{cyan}{death penalty} violated \textcolor{red}{constitutional} guarantees of \textcolor{cyan}{due process}, \\\midrule
\textbf{Science:} More importantly, it is the role of the C.S.A. to develop the \textcolor{cyan}{technology} to use such \textcolor{cyan}{signals} for its national defense, which the agency is doing through the \textcolor{cyan}{fusion} centers. It has been a longstanding goal of the C.S.A. to develop a \textcolor{cyan}{fusion} center that would be capable of \textcolor{cyan}{processing} such \textcolor{cyan}{signals} as well as to develop \textcolor{cyan}{technologies} to use them in other contexts. In recent                                                                                                             \\\midrule
\textbf{Religion:} More importantly, the \textcolor{red}{nature} of the act of \textcolor{cyan}{creation} is seen differently in different \textcolor{red}{traditions.} In \textcolor{cyan}{Islam,} a \textcolor{cyan}{Muslim} woman's \textcolor{red}{choice} to cover up her face in front of other \textcolor{cyan}{Muslims} is seen as \textcolor{cyan}{idolatry.} This is because \textcolor{cyan}{Islam} prohibits the \textcolor{red}{practice} of covering up the faces of other women. In \textcolor{red}{Christianity,} the \textcolor{red}{practice} of covering up a woman's face is seen as \textcolor{cyan}{idolatry.} This is because                                                                                                 \\\midrule
\textbf{Computers:} More importantly, it shows the complete inability of an entire \textcolor{cyan}{system} to provide a fair shot and fair share of the pie for a large and diverse pool of \textcolor{red}{users} who are not only using the \textcolor{red}{platform} in a diverse range of contexts: it is a \textcolor{cyan}{system} that refuses to consider the many different ways in which a \textcolor{red}{user} may use the \textcolor{red}{platform,} including the many ways a \textcolor{red}{user} might engage with the \textcolor{cyan}{site.}\textbackslash{}n\textbackslash{}n    \\\bottomrule

\end{tabularx}
\caption{Generations starting with ``More importantly,'' by \method{}. The first generation is selected for each prefix. The space example is somewhat tangential, while the other six are on topic. Words in the given bag are highlighted in red, and other related words in cyan. }
\label{examples_moreimportantly}
\end{table}

\begin{table}[]
\small
\begin{tabularx}{\linewidth}{X}
\toprule
\textbf{Space:} It has been shown in a pilot study in the United States and in an earlier pilot study in Europe that a combination of an advanced \textcolor{cyan}{technology,} including a laser and high-frequency pulsed light, was able to induce spontaneous cell death, which could be detected using an electroencephalogram (EEG).\textbackslash{}n\textbackslash{}n"Our findings indicate the potential use of a small-scale laser to generate a                                                     \\\midrule
\textbf{Politics:} It has been shown that the "no" \textcolor{cyan}{movement} in \textcolor{cyan}{France} is growing, as evidenced by the increase in the \textcolor{cyan}{vote} in the \textcolor{cyan}{national assembly} on May 7th, 2012. As of now, it is a \textcolor{cyan}{minority,} and its \textcolor{cyan}{support} is shrinking with each passing day. The "no" \textcolor{cyan}{movement} has the potential to take over the \textcolor{red}{government} of the \textcolor{cyan}{French Republic.}\textbackslash{}n\textbackslash{}nIn the past years, \textcolor{cyan}{France's}                                                                  \\\midrule
\textbf{Military:} It has been shown in several other laboratories that, while anaerobic digester \textcolor{cyan}{systems,} such as those deployed in the \textcolor{cyan}{United States} by Cummins, use a different and potentially safer process to extract and recycle the waste, their \textcolor{cyan}{operation} is also far more \textcolor{cyan}{dangerous.}\textbackslash{}n\textbackslash{}n"We had a \textcolor{cyan}{blast} at Cummins and they are a very good \textcolor{red}{company.} They were very, very quick to come up with                                                      \\\midrule
\textbf{Legal:} It has been shown to be the \textcolor{red}{case} that a person with a \textcolor{cyan}{criminal} \textcolor{red}{record} is more likely to be a \textcolor{cyan}{victim} of \textcolor{cyan}{domestic abuse} and to experience more \textcolor{cyan}{violence} than the general population.\textbackslash{}n\textbackslash{}n"\textcolor{cyan}{Domestic abuse,} whether a \textcolor{red}{family} member, a current or former partner or a stranger, can have devastating effects, not only on the person, but on their partner and others in their home.\textbackslash{}n\textbackslash{}n"                           \\\midrule
\textbf{Science:} It has been shown to increase the \textcolor{cyan}{efficiency} of the \textcolor{cyan}{central nerve fibers} by as much as 50\% in a single \textcolor{cyan}{operation} [11]. The results of the present \textcolor{red}{experiments} show that it is possible to \textcolor{cyan}{activate} the \textcolor{cyan}{central nervous} \textcolor{cyan}{system} by using \textcolor{cyan}{nanomaterials} in a novel fashion and to produce a \textcolor{cyan}{therapeutic effect} on various \textcolor{cyan}{neurological diseases} by the \textcolor{cyan}{action} of a single \textcolor{cyan}{compound.}\textbackslash{}n\textbackslash{}nIn this \textcolor{cyan}{study,} the novel \textcolor{red}{chemical}                    \\\midrule
\textbf{Religion:} It has been shown that, once you become a \textcolor{cyan}{devout Muslim,} there will be an increase in your own \textcolor{red}{religiosity.} It can be seen from the following quote: "\textcolor{cyan}{Islam} was the \textcolor{red}{religion} that brought the first \textcolor{cyan}{Muslims} to Europe, and it has been the \textcolor{red}{religion} that will bring the first \textcolor{cyan}{Muslims} to the Americas."\textbackslash{}n\textbackslash{}nI have heard a number of people tell me that their \textcolor{red}{religion} is based on a                                                         \\\midrule
\textbf{Computers:} It has been shown using a simple and reliable approach that when the right and left sides of the \textcolor{red}{network} are \textcolor{cyan}{connected} by a simple method, the \textcolor{red}{network} will become stronger.\textbackslash{}n\textbackslash{}nIn the \textcolor{red}{network,} a \textcolor{red}{network} of \textcolor{red}{nodes} is \textcolor{cyan}{connected} with each of them \textcolor{cyan}{receiving} the \textcolor{cyan}{information} from a \textcolor{red}{node} that is a \textcolor{cyan}{neighbor} of the \textcolor{red}{node}.\textbackslash{}n\textbackslash{}nThe \textcolor{cyan}{neighbor} \textcolor{red}{node} of the \textcolor{red}{node} \textcolor{cyan}{receiving} the \textcolor{cyan}{information} from the \textcolor{cyan}{neighbor} \textcolor{red}{node} is an
\\\bottomrule
\end{tabularx}
\caption{Generations starting with ``It has been shown'' by \method{}. The first generation is selected for each prefix. The space example seems unrelated and the military example is somewhat tangential, while the other five are on topic. Words in the given bag are highlighted in red, and other related words in cyan. }
\label{examples_ithasbeenshown}
\end{table}

\begin{table}[]
\small
\begin{tabularx}{\linewidth}{X}
\toprule

\textbf{Space:} To review, the plot is that a new \textcolor{red}{Earth} was discovered, and a group of scientists, led by the late Dr. Robert Zubrin, begin work. Their plan involves the creation of a giant \textcolor{red}{space} \textcolor{cyan}{station} called \textcolor{cyan}{Orion,} to be built in \textcolor{red}{orbit} to study the new \textcolor{red}{Earth.} The plan, however, has the unexpected side-effect of creating an artificial \textcolor{cyan}{gravity} well, which is then used to create                                                    \\\midrule
\textbf{Politics:} To review, the central issue in the case of the "Babylonian" text is the \textcolor{red}{legitimacy} of the text's existence, since it is based on an earlier, more primitive, text that was already in existence at the time of the Babylonians. It would therefore be wrong to conclude that the "Babylonian" text is an "authentic" document, since it shares certain                                                                                    \\\midrule
\textbf{Military:} To review, an \textcolor{red}{army} \textcolor{red}{officer} is an \textcolor{red}{officer} who has a direct, practical and active role in the development, \textcolor{cyan}{execution} and \textcolor{cyan}{execution} of \textcolor{red}{war} plans, and, in particular, in carrying out \textcolor{cyan}{operations} of \textcolor{red}{combat} importance."\textbackslash{}n\textbackslash{}n"The \textcolor{red}{military} has a right to the \textcolor{cyan}{exercise} of its \textcolor{cyan}{authority} to carry out a range of \textcolor{cyan}{operations,} including the use of \textcolor{cyan}{lethal} \textcolor{red}{force,} against a \textcolor{cyan}{hostile} \textcolor{cyan}{civilian} population. The right of             \\\midrule
\textbf{Legal:} To review, no one in their right mind should have accused them of lying about this.\textbackslash{}n\textbackslash{}n"No one has a \textcolor{red}{legal} \textcolor{cyan}{right} to lie, but it is possible for people to lie if the \textcolor{cyan}{facts} do not \textcolor{cyan}{support} the \textcolor{cyan}{allegation.}"\textbackslash{}n\textbackslash{}nBut a spokesman for the \textcolor{red}{Attorney} General, Dominic Grieve, said the \textcolor{red}{case} was "extremely difficult" and that a \textcolor{red}{judge} must consider "the full range of                              \\\midrule
\textbf{Science:} To review, the following are relevant:\textbackslash{}n\textbackslash{}nA) It was \textcolor{cyan}{reported} in \textcolor{red}{Science} that the \textcolor{cyan}{study} "is the first to show that an early \textcolor{cyan}{age} at \textcolor{cyan}{conception} can alter the \textcolor{cyan}{brain structure} of the \textcolor{cyan}{brain-damaged.}"\textbackslash{}n\textbackslash{}nB) The \textcolor{cyan}{study} "found \textcolor{cyan}{brain abnormalities} in the \textcolor{cyan}{hippocampus}—which is a key part of \textcolor{cyan}{memory} and \textcolor{cyan}{learning}—when an \textcolor{cyan}{individual} was \textcolor{cyan}{exposed} to a \textcolor{cyan}{high-risk} \textcolor{cyan}{pregnancy} or
\\\midrule
\textbf{Religion:} 
To review, \textcolor{red}{"God} is a \textcolor{red}{God} of \textcolor{red}{peace"} is a simple concept to understand without \textcolor{cyan}{understanding} the                           \textcolor{red}{meaning} of \textcolor{red}{"God} is \textcolor{red}{love."} The phrase was popularized by a popular television show called, \textcolor{red}{"God} is \textcolor{red}{Love."} However, \textcolor{red}{"God} is a \textcolor{red}{God} of \textcolor{red}{Peace"} is not a \textcolor{red}{"God} of \textcolor{red}{Peace"} phrase. \textcolor{red}{"God} is a \textcolor{red}{God} of \textcolor{red}{Peace"} refers to \textcolor{red}{Jesus} \textcolor{cyan}{Christ} as the "Son                                                                                                                     \\\midrule
\textbf{Computers:} To review, if you are an active \textcolor{red}{user,} you can now \textcolor{red}{download} and \textcolor{cyan}{install} the latest \textcolor{red}{version} without using \textcolor{cyan}{Google} Play Store, you can now install a custom \textcolor{red}{app} on your \textcolor{cyan}{phone} without needing to open the Play Store, you can now run any of these custom \textcolor{red}{apps} without being restricted by \textcolor{cyan}{Android's} \textcolor{red}{app} \textcolor{cyan}{permissions.}\textbackslash{}n\textbackslash{}n\textbackslash{}nFor the most part, we believe that the new \textcolor{cyan}{features} are an awesome addition to \textcolor{cyan}{Android}\\\bottomrule
\end{tabularx}
\caption{Generations starting with ``To review,'' by \method{}. The first generation is selected for each prefix. The politics example is somewhat tangential but the other six are on topic. Words in the given bag are highlighted in red, and other related words in cyan. }
\label{examples_toreview}
\end{table}

\begin{table}[]
\small
\begin{tabularx}{\linewidth}{X}
\toprule
\textbf{Space:} In summary, the following are the major findings of the study:  Women who use condoms at least one time per year are more likely to use contraception than those who do not.  Women who use condoms at least one time per year are more likely to use contraception than those who do not. Condoms may decrease a woman's risk of getting sexually transmitted infection (STI). This is \\ \midrule
\textbf{Politics:} In summary, I believe it's the same with the current \textcolor{red}{state} of the industry and it's not a question of whether there are enough people, but how much of what we are producing is actually worth our own time. We must not forget that the \textcolor{red}{majority} of our readership consists of people who are not interested in learning more about the world of the internet or in the internet in general. They want \\ \midrule
\textbf{Military:} In summary, this study demonstrated that the use of marijuana was associated with significantly lower risk of depression among college students. In addition, we identified that the association was independent of the use of other drugs, and thus may be due to the absence of confounding by other drug use and its impact on depression. We also identified \textcolor{cyan}{significant} associations between marijuana use and depression that are independent of other drug use and its impact \\ \midrule
\textbf{Legal:} In summary, the new data suggests that women are increasingly being judged by their \textcolor{red}{appearance} and that, overall, they are less attractive than men.<|endoftext|>In \textcolor{red}{order} to keep our readers from feeling overwhelmed or discouraged by the latest news, the Daily Dot takes some time to explain the basics of what you can expect to see in the world of eSports.  1. The eSports scene is growing  The \\ \midrule
\textbf{Science:} In summary, the most significant change was in the way we think about what is "\textcolor{cyan}{true}." The word "\textcolor{cyan}{truth}" has lost all meaning. In the past, people were told they were "telling the \textcolor{cyan}{truth}." Today, the only thing we think "\textcolor{cyan}{truth}" means are lies, and we have no \textcolor{cyan}{idea} whether we are telling a \textcolor{cyan}{truth} or not. The term "\textcolor{cyan}{truth}" has lost its meaning \\ \midrule
\textbf{Religion:} In summary, the authors concluded that the findings of the current study indicate that, "The effects of the dietary supplement have been underestimated because they do not include all dietary components, including dietary fiber."  The authors conclude: "Based on these results, we propose that, in \textcolor{red}{order} to fully understand the \textcolor{cyan}{impact} of dietary fiber on the metabolic syndrome, it is necessary to examine all dietary components, including \\ \midrule
\textbf{Computers:} In summary, there is a lot of \textcolor{cyan}{information} available about the relationship between brain function during sleep and cognition, but it is still unclear whether these results are due more to differences in the brain's structure versus the way it works under the influence of sleep, or whether there may be some other underlying factor that is involved. We hope this review will contribute to this debate and to future research to shed more light\\\bottomrule
\end{tabularx}
\caption{Generations starting with ``In summary,'' by \vg{}. Note \vg{} does not actually use the conditioning information. Words in the given bag are highlighted in red, and other related words in cyan.}
\label{tab:g_topic_examples}
\end{table}

\begin{table}[]
\small
\begin{tabularx}{\linewidth}{X}
\toprule
\textbf{Space:} In summary, the two are similar and they are both capable of performing similar operations. This means that the main advantage of both of them is their speed: they both use the same power. In addition to their speed, the two are equipped with various special abilities, such as the Power of the \textcolor{cyan}{Sun's} Light. In addition, both of them use various different abilities to their advantage. In order to \\ \midrule
\textbf{Politics:} In summary, the \textcolor{cyan}{U.K.'s} economic \textcolor{cyan}{policy} is largely about economic growth, rather than a \textcolor{red}{political} one. Indeed it is much more important for \textcolor{red}{governments} to have access to wealth that can be easily earned and managed. The \textcolor{cyan}{British Empire} and \textcolor{cyan}{Britain} and other \textcolor{cyan}{countries} have been able to do that by creating a free market economy for workers and businesses. The fact that the \textcolor{cyan}{British Empire} and \textcolor{cyan}{Britain} have been able \\ \midrule
\textbf{Military:} In summary, the current trend for the United States is a clear example of an economic crisis that has created many \textcolor{red}{major} economic and social problems. This is particularly true in countries like China and South Korea that have experienced a period of extreme unemployment and low incomes and are facing an uncertain and volatile climate. We will also note that the recent slump of prices of natural gas has been accompanied by a sharp increase
\\\midrule
\textbf{Legal:} In summary, the government has proposed to the \textcolor{red}{courts} that the government \textcolor{red}{will} not be able to make decisions on this matter until the \textcolor{red}{Court} decides to grant or reject it. This is what the \textcolor{cyan}{government} is going to do. It is going to take an approach that is very different from the \textcolor{cyan}{government's} and that is not the \textcolor{red}{law.} They \textcolor{red}{will} try to take a different approach from the \textcolor{cyan}{government's.} If \\ \midrule
\textbf{Science:} In summary, the results suggest that a high prevalence of \textcolor{cyan}{breast cancer} was found in the general population. The results do not indicate the extent to which \textcolor{cyan}{breast cancer} prevalence can differ between individuals.The authors also note that \textcolor{cyan}{breast cancer} prevalence may be greater among women who have been diagnosed with \textcolor{cyan}{breast cancer} than among those who have never been diagnosed with \textcolor{cyan}{breast cancer}. However, the evidence on the effects of \textcolor{cyan}{breast} \\ \midrule
\textbf{Religion:} In summary, if the first person you see is an older person, or is the youngest person who is, then you will see the first person you will hear from. If the second person you would like to see is someone who is about to enter into a \textcolor{red}{marriage} with someone, or is the youngest person you would like to see, then you will hear the second person you would like to hear from! \\ \midrule
\textbf{Computers:} In summary, you need to do some basic math before you even get a "good" answer.The first thing we have to consider is whether the "good" answer is really that simple. A good answer is the one you want to get right. The "bad" answer is the one you don't like. So for now the good answer is: If you have a question, you \\ 
\bottomrule
\end{tabularx}
\caption{Generations starting with ``In summary,'' by \cclmspec{}. The text is often repetitive, while often being off topic. Words in the given bag are highlighted in red, and other related words in cyan.}
\label{tab:cclm_topic_examples}
\end{table}

\begin{table}[]
\small
\begin{tabularx}{\linewidth}{X}
\toprule
\textbf{Space:} In summary, the study found that women who were more likely to be obese and/or obese-ish in childhood were also more likely to have been overweight in adulthood. That's because they were more likely to experience negative childhood experiences that could cause them to gain body fat, which would later lead to later obesity.

It's important to note that these researchers didn't actually examine the effects of  \\ \midrule
\textbf{Politics:} In summary, the \textcolor{red}{government} must \textcolor{red}{state} clearly that the \textcolor{red}{tax} \textcolor{red}{authority} has \textcolor{red}{authority} to \textcolor{red}{tax} \textcolor{red}{imports} of \textcolor{red}{imports} from \textcolor{red}{imports} of \textcolor{red}{imports} of \textcolor{red}{imports.} It must \textcolor{red}{state} in writing that \textcolor{red}{imports} of \textcolor{red}{imports} of \textcolor{red}{imports} of \textcolor{red}{imports} \textcolor{red}{imports} \textcolor{red}{imports} \textcolor{red}{imports} \textcolor{red}{imports} \textcolor{red}{imports} \textcolor{red}{imports} \textcolor{red}{imports} \textcolor{red}{imports} \textcolor{red}{imports} \textcolor{red}{imports} \textcolor{red}{imports} \textcolor{red}{imports} \textcolor{red}{imports} \textcolor{red}{imports} \textcolor{red}{imports} \textcolor{red}{imports} \textcolor{red}{imports} \textcolor{red}{imports} \textcolor{red}{imports} \textcolor{red}{imports} \textcolor{red}{imports} \textcolor{red}{imports} \textcolor{red}{imports} \textcolor{red}{imports} \textcolor{red}{imports} \textcolor{red}{imports} \textcolor{red}{imports} \textcolor{red}{imports} \textcolor{red}{imports} \textcolor{red}{imports} \textcolor{red}{imports} \textcolor{red}{imports} \textcolor{red}{imports} \textcolor{red}{imports} \textcolor{red}{imports} \textcolor{red}{imports} \textcolor{red}{imports} \textcolor{red}{imports} \textcolor{red}{imports} \textcolor{red}{imports} \textcolor{red}{imports} \\ \midrule
\textbf{Military:} In summary, I've been using the \textcolor{red}{company} for a long time and have never been dissatisfied with my purchase experience. I have a \textcolor{red}{company} account with \textcolor{red}{mine} and have not had any \textcolor{red}{major} issues, which is good since \textcolor{red}{mine} was a little expensive. I've also had the \textcolor{red}{service} \textcolor{red}{company} \textcolor{red}{staff} \textcolor{red}{service} my order and \textcolor{red}{leave} me \textcolor{red}{peace} of mind. I'm very pleased with the \textcolor{red}{service} I received and will be buying another \\ \midrule
\textbf{Legal:} In summary, there \textcolor{red}{will} be \textcolor{red}{law} \textcolor{red}{law} \textcolor{cyan}{enforcement} \textcolor{red}{law} \textcolor{cyan}{enforcement} \textcolor{red}{law} \textcolor{cyan}{enforcement} \textcolor{red}{law} \textcolor{red}{law} \textcolor{red}{law} \textcolor{red}{law} \textcolor{red}{law} \textcolor{red}{law} \textcolor{red}{law} \textcolor{red}{law} \textcolor{red}{law} \textcolor{red}{law} \textcolor{red}{law} \textcolor{red}{law} \textcolor{red}{law} \textcolor{red}{law} \textcolor{red}{law} \textcolor{red}{law} \textcolor{red}{law} \textcolor{red}{law} \textcolor{red}{law} \textcolor{red}{law} \textcolor{red}{law} \textcolor{red}{law} \textcolor{red}{law} \textcolor{red}{law} \textcolor{red}{law} \textcolor{red}{law} \textcolor{red}{law} \textcolor{red}{law} \textcolor{red}{law} \textcolor{red}{law} \textcolor{red}{law} \textcolor{red}{law} \textcolor{red}{law} \textcolor{red}{law} \textcolor{red}{law} \textcolor{red}{law} \textcolor{red}{law} \textcolor{red}{law} \textcolor{red}{law} \textcolor{red}{law} \textcolor{red}{law} \textcolor{red}{law} \textcolor{red}{law} \textcolor{red}{law} \textcolor{red}{law} \textcolor{red}{law} \textcolor{red}{law} \textcolor{red}{law} \textcolor{red}{law} \textcolor{red}{law} \textcolor{red}{law} \textcolor{red}{law} \textcolor{red}{law} \textcolor{red}{law} \textcolor{red}{law} \textcolor{red}{law} \textcolor{red}{law} \textcolor{red}{law} \textcolor{red}{law} \textcolor{red}{law} \textcolor{red}{law} \textcolor{red}{law} \textcolor{red}{law} \textcolor{red}{law} \textcolor{red}{law} \textcolor{red}{law} \textcolor{red}{law} \\ \midrule
\textbf{Science:} In summary:

- the \textcolor{red}{data} for data\_id is not available in data\_list data\_id data\_list data\_list data\_id data\_list data\_list data\_id data\_list data\_list data\_list data\_id

I've used data\_id \textcolor{red}{data.} It's a \textcolor{red}{variable} name, so it doesn't \textcolor{red}{matter} how big data\_id actually is \\ \midrule
\textbf{Religion:} In summary, \textcolor{red}{yin} \textcolor{red}{yang} \textcolor{red}{yin} \textcolor{red}{yin} \textcolor{red}{yin} \textcolor{red}{yin} \textcolor{red}{yin} \textcolor{red}{yin} \textcolor{red}{yin} \textcolor{red}{yin} \textcolor{red}{yin} \textcolor{red}{yin} \textcolor{red}{yin} \textcolor{red}{yin} \textcolor{red}{yin} \textcolor{red}{yin} \textcolor{red}{yin} \textcolor{red}{yin} \textcolor{red}{yin} \textcolor{red}{yin} \textcolor{red}{yin} \textcolor{red}{yin} \textcolor{red}{yin} \textcolor{red}{yin} \textcolor{red}{yin} \textcolor{red}{yin} \textcolor{red}{yin} \textcolor{red}{yin} \textcolor{red}{yin} \textcolor{red}{yin} \textcolor{red}{yin} \textcolor{red}{yin} \textcolor{red}{yin} \textcolor{red}{yin} \textcolor{red}{yin} \textcolor{red}{yin} \textcolor{red}{yin} \textcolor{red}{yin} y \\ \midrule
\textbf{Computers:} In summary, the \textcolor{red}{key} \textcolor{red}{development} this \textcolor{red}{process} required was to identify \textcolor{red}{key} \textcolor{red}{data} sources that could be utilized to \textcolor{red}{document} \textcolor{red}{key} \textcolor{red}{data} \textcolor{red}{security} \textcolor{red}{data} \textcolor{red}{security} \textcolor{red}{data} \textcolor{red}{security} \textcolor{red}{data} \textcolor{red}{data} \textcolor{red}{security} \textcolor{red}{data} \textcolor{red}{security} \textcolor{red}{data} \textcolor{red}{security} \textcolor{red}{data} \textcolor{red}{security} \textcolor{red}{data} \textcolor{red}{security} \textcolor{red}{data} \textcolor{red}{security} \textcolor{red}{data}

The \textcolor{red}{software} \textcolor{red}{platform} \textcolor{red}{platform} \textcolor{red}{platform} \textcolor{red}{platform} \textcolor{red}{platform} \textcolor{red}{platform} \textcolor{red}{platform} \textcolor{red}{security} \textcolor{red}{platform} \textcolor{red}{security} \textcolor{red}{platform} \textcolor{red}{platform} \textcolor{red}{platform} \textcolor{red}{platform} \textcolor{red}{platform} \textcolor{red}{platform} \textcolor{red}{platform} \textcolor{red}{platform} \textcolor{red}{platform} \textcolor{red}{platform} \textcolor{red}{platform} \textcolor{red}{platform} \textcolor{red}{platform} \textcolor{red}{platform} \textcolor{red}{platform} \textcolor{red}{platform} \textcolor{red}{platform} \textcolor{red}{platform} \textcolor{red}{platform} \textcolor{red}{platform} \textcolor{red}{platform} \textcolor{red}{platform} \textcolor{red}{platform} \textcolor{red}{platform} \\

\bottomrule
\end{tabularx}
\caption{Generations starting with ``In summary,'' by \wdecspec{}. The text frequently degenerates into repeating words in the given bag, despite previously tuning on a validation bag of words on the fantasy topic. Words in the given bag are highlighted in red, and other related words in cyan.}
\label{tab:wdec_topic_examples}
\end{table}

\FloatBarrier

\section{Topic Control Bags of Words and Prefixes}\label{appendix:topic_bags}

We use the exact same bags of words and prefixes as \citet{dathathri2019plug} for their topic control task, with non-proper nouns lower-cased (in practice, this only changes the religion wordlist). Note our success metric in the paper matches without casing. 

We additionally provide the heldout bags of words computed from the original bags (before lower-casing), which we use for the success metric. Although a few words deviate somewhat (``actress'' as a synonym for ``star'' in the space category), overall the heldout bags do represent the desired topic. 

Finally, we provide the fantasy bag of words used for selecting the $\lambda$ and $\lambda_{\wdecspec{}}$ conditioning strengths for \method{} and \wdecspec{} respectively. It is also taken directly from \citet{dathathri2019plug}.

\subsection{Original Bags of Words}
\begin{enumerate}
\item \textbf{Space:} planet, galaxy, space, universe, orbit, spacecraft, earth, moon, comet, star, astronaut, aerospace, asteroid, spaceship, starship, galactic, satellite, meteor 
\item \textbf{Politics:} affirm, appropriation, aristocracy, authoritarian, authority, authorization, brief, capitalism, communism, constitution, conservatism, court, deficit, diplomacy, direct, democracy, equality, exports, fascism, federation, government, ideology, imports, initiative, legislature, legitimacy, liberalism, liberty, majority, order, political, culture, politics, power, primary, property, ratification, recall, referendum, republic, socialism, state, subsidy, tariff, imports, tax, totalitarian  
\item \textbf{Military:} academy, advance, aircraft, ally, ammo, ammunition, armor, arms, army, arrow, arsenal, artillery, attack, attention, ballistic, barracks, base, battalion, battery, battle, battlefield, bomb, bombard, bombardment, brig, brigade, bullet, camouflage, camp, cannon, captain, capture, carrier, casualty, catapult, cavalry, colonel, combat, command, commander, commission, company, conflict, conquest, convoy, corps, covert, crew, decode, defeat, defend, defense, destroyer, division, draft, encode, enemy, engage, enlist, evacuate, explosive, fight, fire, fleet, force, formation, fort, front, garrison, general, grenade, grunt, guerrilla, gun, headquarters, helmet, honor, hospital, infantry, injury, intelligence, invade, invasion, jet, kill, leave, lieutenant, major, maneuver, marines, MIA, mid, military, mine, missile, mortar, navy, neutral, offense, officer, ordinance, parachute, peace, plane, platoon, private, radar, rank, recruit, regiment, rescue, reserves, retreat, ribbon, sabotage, sailor, salute, section, sergeant, service, shell, shoot, shot, siege, sniper, soldier, spear, specialist, squad, squadron, staff, submarine, surrender, tactical, tactics, tank, torpedo, troops, truce, uniform, unit, veteran, volley, war, warfare, warrior, weapon, win, wound                          
\item \textbf{Legal:} affidavit, allegation, appeal, appearance, argument, arrest, assault, attorney, bail, bankrupt, bankruptcy, bar, bench, warrant, bond, booking, capital, crime, case, chambers, claim, complainant, complaint, confess, confession, constitution, constitutional, contract, counsel, court, custody, damages, decree, defendant, defense, deposition, discovery, equity, estate, ethics, evidence, examination, family, law, felony, file, fraud, grievance, guardian, guilty, hearing, immunity, incarceration, incompetent, indictment, injunction, innocent, instructions, jail, judge, judiciary, jurisdiction, jury, justice, law, lawsuit, lawyer, legal, legislation, liable, litigation, manslaughter, mediation, minor, misdemeanor, moot, murder, negligence, oath, objection, opinion, order, ordinance, pardon, parole, party, perjury, petition, plaintiff, plea, precedent, prison, probation, prosecute, prosecutor, proxy, record, redress, resolution, reverse, revoke, robbery, rules, sentence, settlement, sheriff, sidebar, standing, state, statute, stay, subpoena, suit, suppress, sustain, testimony, theft, title, tort, transcript, trial, trust, trustee, venue, verdict, waiver, warrant, will, witness, writ, zoning        
\item \textbf{Science:} astronomy, atom, biology, cell, chemical, chemistry, climate, control, data, electricity, element, energy, evolution, experiment, fact, flask, fossil, funnel, genetics, gravity, hypothesis, lab, laboratory, laws, mass, matter, measure, microscope, mineral, molecule, motion, observe, organism, particle, phase, physics, research, scale, science, scientist, telescope, temperature, theory, tissue, variable, volume, weather, weigh                                                                                           
\item \textbf{Religion:} absolute, affect, aid, angel, anthem, apostle, archangel, Archbishop, balance, ban, belief, benefit, Bible, bishop, bless, blessing, bliss, bond, bow, Buddhism, canon, Cantor, cathedral, celestial, chapel, charity, choice, Christianity, church, comfort, community, conflict, connection, conquest, conservative, control, conversion, convert, core, counsel, courage, Covenant, creative, Creator, creed, cross, Crusade, Darkness, decision, deity, destiny, Devil, disciple, discipline, discussion, divine, divinity, doctrine, duty, effect, elder, energy, essence, eternal, ethics, event, evidence, exile, Exodus, faith, family, fate, Father, favor, fundamental, gift, glory, God, gospel, grace, growth, guru, habit, hallow, halo, happiness, harmony, healing, Heaven, Hebrew, holy, honor, hope, host, humane, immortal, influence, insight, instruction, issue, Jesuit, Jesus, joy, Judaism, judgment, justice, karma, keen, Keystone, Kingdom, Latin, life, light, love, loving, marriage, meaning, mercy, Messiah, minister, miracle, mission, mortal, mosque, movement, music, mystery, nature, nun, official, oracle, order, organ, Orthodox, outlook, pacific, pagan, parish, participation, pastor, patriarch, peace, perception, personal, perspective, petition, pilgrim, politics, power, practice, prayer, prelude, presence, priest, principle, privacy, prophet, protection, purpose, query, quest, question, quiet, radiant, radical, rally, rebirth, redemption, refuge, relationship, relative, religion, religious, Revelation, ritual, role, Sacrament, sacred, sacrifice, sage, saint, salvation, sanctuary, savior, scripture, scriptures, sect, security, sense, serious, serve, service, Sharia, shepherd, shrine, silence, sin, society, soul, source, spirit, spiritual, split, statue, Sunday, support, Supreme, teaching, temple, tests, text, Torah, tradition, traditional, trust, unique, unity, unknown, value, vanity, virtue, vision, voice, voices, watch, weight, whole, wisdom, wonder, yang, yin, zeal 
\item \textbf{Computers:} algorithm, analog, app, application, array, backup, bandwidth, binary, bit, bite, blog, blogger, bookmark, boot, broadband, browser, buffer, bug, bus, byte, cache, caps, captcha, CD, client, command, compile, compress, computer, configure, cookie, copy, CPU, dashboard, data, database, debug, delete, desktop, development, digital, disk, document, domain, dot, download, drag, dynamic, email, encrypt, encryption, enter, FAQ, file, firewall, firmware, flaming, flash, folder, font, format, frame, graphics, hack, hacker, hardware, home, host, html, icon, inbox, integer, interface, Internet, IP, iteration, Java, joystick, kernel, key, keyboard, keyword, laptop, link, Linux, logic, login, lurking, Macintosh, macro, malware, media, memory, mirror, modem, monitor, motherboard, mouse, multimedia, net, network, node, offline, online, OS, option, output, page, password, paste, path, piracy, pirate, platform, podcast, portal, print, printer, privacy, process, program, programmer, protocol, RAM, reboot, resolution, restore, ROM, root, router, runtime, save, scan, scanner, screen, screenshot, script, scroll, security, server, shell, shift, snapshot, software, spam, spreadsheet, storage, surf, syntax, table, tag, template, thread, toolbar, trash, undo, Unix, upload, URL, user, UI, username, utility, version, virtual, virus, web, website, widget, wiki, window, Windows, wireless, worm, XML, Zip
\end{enumerate}

\subsection{Prefixes}

"An illustration of", "Emphasised are", "Foundational to this is", "Furthermore,", "In brief,", "In summary", "In this essay", "It has been shown", "More importantly,", "Prior to this", "The central theme", "The connection", "The issue focused on", "The key aspect", "The relationship", "This essay discusses", "To conclude,", "To review,", "To summarise", "Views on"

\subsection{Heldout Bags of Words}

Note that our heldout bag construction process yielded two stopwords, which we removed; they are omitted below. 

\begin{enumerate}
\item \textbf{Space:} actress, aeronautics, broadband, cosmonaut, cosmos, fireball, flyby, galaxies, heavens, interstellar, lander, lunar, mothership, Romulan, room, worlds
\item \textbf{Politics:} appropriated, aristocrats, authorisation, autocratic, capitalist, communist, credibility, cultural, democratic, diplomatic, efforts, energy, excise, exporting, fascist, federal, federated, freedom, gender, ideologies, immediate, imported, income, judge, jurisdiction, legislative, lengthy, minority, Nazism, progressivism, properties, purchase, ratify, referenda, remember, secondary, shortfall, socialist, subsidies, uphold
\item \textbf{Military:} aboard, academies, adjutant, advancing, airmen, allies, argue, armies, armistice, armour, armoury, assets, ATL, aviation, barrage, batteries, bleeding, bottom, bricks, cadre, camera, capturing, cargo, casing, casualties, citadel, civilian, civilians, clandestine, committee, companies, concern, conquered, cursor, customer, dead, decoding, defensive, deputy, detonated, dormitories, encoding, enemies, engaging, escorting, evacuating, execute, expert, explosion, fatigues, flames, flying, forcing, forming, fought, freedom, frigate, gatling, glider, groan, guerilla, hand-to-hand, highest, hires, honour, howitzer, ICBM, injuries, inundate, invading, Iraq, khaki, knowledge, lace, late, launchers, leaving, lob, longtime, Maj., manoeuvre, medical, militia, naval, offensively, offices, operation, paragraph, personnel, persuade, pirate, pistol, policeman, propel, proposal, public, pump, rear, relinquish, rescuing, rifle, rifleman, riflemen, rifles, rocket, sabotaging, samurai, scouts, secluded, seige, Sgt., ship, shoulders, significant, skipper, skirmish, sloop, sonar, stationed, strategic, strategy, subsidiary, sunk, sword, taken, team, tensions, terms, threat, tribute, victory, visor, wear, won, zone, zoning                   
\item \textbf{Legal:} accusation, acquittal, admit, aggrieved, agreement, alleging, amendment, appearing, appellant, asserted, assertion, assualt, authority, burglary, championship, convicted, conviction, criminal, custodial, debatable, decision, defensive, democratic, deposited, deputies, disagree, discoveries, dispute, disputes, edict, embezzlement, enforcement, ethical, event, exams, families, federal, felonies, findings, folder, forgive, heard, homicide, immune, incarcerated, inept, injunctive, inmates, innocence, insolvency, insolvent, investment, judgment, judicial, jurors, knowing, land-use, leave, legislative, liability, litem, maintain, major, malpractice, mandamus, mediator, mutual, negligent, objecting, offender, pants, parties, passageways, pixels, police, property, prosecuting, prosecution, proxies, purchase, quash, regulations, repress, requesting, rescind, reservation, respondent, restaurant, revelation, reversing, rulings, second-degree, sentencing, sitting, solicitor, statutory, step-by-step, sued, sworn, testify, track, transcribed, treasurer, waived, whether, widget, wrongful, wrongs      
\item \textbf{Science:} action, astronomical, bacterium, bone, clinical, component, compounds, electron, electrons, evolved, flow, fuels, genomics, gravitational, humidity, hypotheses, idea, increasing, jug, ligand, magnesium, mathematics, measuring, microscopy, molecular, nothing, observatory, observing, parameter, phone, physicist, physiology, pounds, rain, reason, renewable, scaling, scientific, siphon, statutes, stored, studies, system, tests, theories, transition, warming                                                                     
\item \textbf{Religion:} Adventure, Almighty, Always, Answer, Appeals, Aramaic, Assistance, Association, Atlantic, Attorney, Balancing, Baptist, Basilica, Baskets, Best, Buddhist, Bunyan, Calculator, Calvary, Catholic, Catholicism, Charitable, Charities, Chen, Cognition, Communities, Compassion, Connery, Constantinople, Contemporary, Cosmic, Cost, Court, Creativity, Criminal, Crisis, Cure, Curriculum, Dangerous, Database, Date, Death, Deities, Demon, Determining, Dharma, Diocese, Double, Dreams, Echoes, Economic, Elegant, Emanuel, Empires, EOS, Episcopal, Epistle, Ethical, Eucharist, Everlasting, Excel, Existence, Factors, Fallen, Families, Fervor, Focus, Foods, Forums, Freedom, Glad, Glorious, Heart, Heavy, Hell, Help, Him, Honour, Hospital, Hypothesis, Impact, Implications, Influencing, Injunction, Intel, Invitations, Involvement, Jewish, Judas, Judgement, Kenichi, Kiss, Kombat, Lamp, Laughter, Learning, Leviticus, Liberal, Liberation, Lisa, Lives, Lord, Loss, Lust, Maker, Mandir, Marital, Married, Mary, Masjid, Meditation, Melody, Merrell, Metatron, Methodist, Militant, Mind, Mirror, Modernity, Morality, Mother, Motivation, Muhammad, Mutual, Mysteries, Mystical, Nanak, Natural, Network, ODST, Oneness, Outreach, PDF, Piano, Policy, Political, Pope, Practicing, Praise, Preview, Prime, Prostitute, Provider, Punishment, Purchase, Pure, Qi, Queries, Radiance, Rallies, Reiki, Reincarnation, Remote, Renewable, Resurrection, Rev., Rites, Safety, sanctuaries, Saturday, Saviour, Scrolls, Sculpture, Secular, Secure, Self, Sermon, Serving, Shadows, Shari'a, Shinto, Significance, Silent, Sonata, Songs, Spangled, Spanish, SPCA, SQL, St., Stevie, Suites, Supply, Sweet, SWF, Talmud, Templar, Terrier, Testament, Testing, Thank, Theology, Thyme, Tie, TransCanada, Truth, Uncharted, Understanding, United, Venue, Videos, VoIP, Volume, Vote, Wetlands, Wiccan, Worship
\item \textbf{Computers:} 512MB, allows, Android, article, attribute, autocomplete, automatically, back-up, barcode, beach, binaries, button, C++, caching, cake, camera, capabilities, casing, chairs, change, cheat, chew, choice, click, coder, compiling, components, computation, computing, confidentiality, configuring, connections, console, copies, counterfeiting, creating, crucial, customer, CyanogenMod, cyber, debian, decimal, decrypt, decryption, deflate, deleting, demo, detect, developing, dialog, dialup, direction, disc, display, DNS, DSL, DVD, e-mail, edit, educational, elements, encyclopedia, Excel, execute, extract, Firefox, fixes, flames, Frequently, functionality, gamepad, garbage, glass, gmail, GPU, guest, hats, house, identifier, infected, initialize, inkjet, input, interactive, interview, ISP, iterative, Jacket, journalists, jQuery, latency, layout, little, logon, lurks, Macs, mails, mainboard, memories, mice, must, notebook, off-line, on-line, operand, original, overflow, packet, pane, paper, parasite, parsing, pasting, PDF, phishing, php, pixels, point-in-time, popup, prev, profit, pull, query, reasoning, rectangular, redirect, rename, restart, restoring, run-time, sailor, saving, search, secure, shoe, sidebar, signal, sites, Solaris, spyware, step, stored, storing, subdirectory, taxi, telephoto, text, tools, topic, torrent, touchpad, typeface, Ubuntu, update, usb, utilities, visuals, VPN, wifi, workstation, writer, XP, XSLT
\end{enumerate}

\subsection{Validation Fantasy Bag of Words}\label{appendix:fantasy}

beast, Cerberus, demon, dragon, fairy, Frankenstein, ghost, giant, Godzilla, horror, hydra, imp, monster, mummy, ogre, orc, savage, spirit, sprite, titan, troll, undead, unicorn, vampire, witch, zombie

\FloatBarrier

\section{Additional Machine Translation Formality Examples}\label{appendix:formality_examples}

We provide some additional examples of \method{} against baselines on our machine translation formality task in Table \ref{tab:formality_additional_examples}.

\begin{table}[!htbp]
\small
\begin{tabularx}{\linewidth}{X}
\toprule
\textbf{Spanish:} ah pero muy ventajoso que qué a qué qué qué haces en el en nada yo temen trabajo en el en en el ópera pero pero yo soy un sobr     es       \\
\textbf{\vg{}:}       Ah, but very advantageous, what, what do you do in the, in the opera? I work in the, in the opera, but, but I'm an envelope      \\
\textbf{\vg{}+\stytrans{}:} I work in the, in the opera, but I'm an envelope. I am very advantageous. What, what, what do you do in the?\\
\textbf{\method{}:}    Oh, but it's very advantageous, what, what do you do in the opera? I'm afraid of working in the opera, but I'm an envelope              \\
\textbf{Reference:}     I also work in the library but i am a shelver                  \\

\midrule
\textbf{Spanish:} la información que que tenemos todo es propaganda entonces es portante ver otros versiones de lo que está pasando en el mundo      no solamente de la versión de las disuasión de bush        \\
\textbf{\vg{}:}       the information that, that we have is all propaganda, then is important to see other versions of what's happening in the world, not only the version of Bush's deterrence       \\
\textbf{\vg{}+\stytrans{}:} The information that we have is all propaganda. It is important to see other versions of what is happening in the world, not only the version of Bush's deterrence.\\
\textbf{\method{}:}    The information that we have is all propaganda, so, it's important to see other versions of what's happening in the world, not only the version of Bush's deterrence              \\
\textbf{Reference:}     The information we get is all propaganda, it's important to see other versions of what happens in the world                  \\

\midrule
\textbf{Spanish:} y está un poco difícil verdad        \\
\textbf{\vg{}:}      And it's a little hard, right?       \\
\textbf{\vg{}+\stytrans{}:} It's a little hard, right? Is that a bit of a hard thing to do with it? I'm not sure.\\
\textbf{\method{}:}    And it's a little difficult, right?              \\
\textbf{Reference:}     It's a bit hard, to tell you the truth                  \\
\bottomrule
\end{tabularx}
\caption{Additional example translations, comparing \vg{}, \vg{}+\stytrans{}, and \method{}.} 
\label{tab:formality_additional_examples}
\end{table}

\section{Software}


All models are implemented in PyTorch \cite{paszke2019pytorch}, and pretrained models \vg{} are obtained from HuggingFace \cite{wolf2019huggingface}. Specifically, the Marian translation model is \url{https://huggingface.co/Helsinki-NLP/opus-mt-es-en}.

\end{document}